\documentclass{article}
\pdfoutput=1

\usepackage[preprint]{neurips_2020}

\usepackage[utf8]{inputenc} 
\usepackage[T1]{fontenc}    
\usepackage{hyperref}       
\usepackage{url}            
\usepackage{booktabs}       
\usepackage{amsfonts}       
\usepackage{nicefrac}       
\usepackage{microtype}      

\usepackage{graphicx, subcaption} 
\usepackage{amsmath} 
\usepackage{amssymb}  
\usepackage{algorithm}
\usepackage[noend]{algpseudocode}

\usepackage{multirow}
\usepackage{lscape}
\usepackage{bbm}
\usepackage{amsmath}

\usepackage{textcomp}

\renewcommand{\textuparrow}{$\uparrow$}
\renewcommand{\textdownarrow}{$\downarrow$}

\graphicspath{ {figures/} }

\DeclareMathOperator{\E}{\mathbb{E}}

\title{
Learning predictive representations in autonomous driving to improve deep reinforcement learning
}

\author{
  Daniel Graves \\
  Noah's Ark Lab \\
  Huawei Technologies Canada, Ltd \\
  Edmonton, Canada \\
  \texttt{daniel.graves@huawei.com}
  \And
  Nhat M. Nguyen \\
  Noah's Ark Lab \\
  Huawei Technologies Canada, Ltd \\
  Edmonton, Canada \\
  \texttt{minh.nhat.nguyen@huawei.com}
  \And
  Kimia Hassanzadeh \\
  Noah's Ark Lab \\
  Huawei Technologies Canada, Ltd \\
  Edmonton, Canada \\
  \texttt{kimia.hassanzadeh@huawei.com}
  \And
  Jun Jin \\
  Noah's Ark Lab \\
  Huawei Technologies Canada, Ltd \\
  Edmonton, Canada \\
  \texttt{jun.jin1@huawei.com}
}

\begin{document}
\maketitle

\begin{abstract}
Reinforcement learning using a novel predictive representation is applied to autonomous driving to accomplish the task of driving between lane markings where substantial benefits in performance and generalization are observed on unseen test roads in both simulation and on a real Jackal robot.
The novel predictive representation is learned by general value functions (GVFs) to provide out-of-policy, or counter-factual, predictions of future lane centeredness and road angle that form a compact representation of the state of the agent improving learning in both online and offline reinforcement learning to learn to drive an autonomous vehicle with methods that generalizes well to roads not in the training data.
Experiments in both simulation and the real-world demonstrate that predictive representations in reinforcement learning improve learning efficiency, smoothness of control and generalization to roads that the agent was never shown during training, including damaged lane markings.
It was found that learning a predictive representation that consists of several predictions over different time scales, or discount factors, improves the performance and smoothness of the control substantially.
The Jackal robot was trained in a two step process where the predictive representation is learned first followed by a batch reinforcement learning algorithm (BCQ) from data collected through both automated and human-guided exploration in the environment.
We conclude that out-of-policy predictive representations with GVFs offer reinforcement learning many benefits in real-world problems.

\end{abstract}

\section{Introduction}
Learning good representations in deep reinforcement learning that generalize to the complexity  encountered by agents in the real-world is a challenging problem \citep{awhite2015}.
Being able to learn and adapt continually by collecting experience is a long standing goal of AI \citep{sutton2011}\citep{modayil2012}.
Agents need to learn and understand the patterns observed in their sensorimotor streams to understand the world \citep{sutton2011}\citep{clark2013}.
This kind of representation is often not focused on solving a particular problem but rather expanding its knowledge and skills to enable the agent to adapt quickly to new problems \citep{schaul2013}.
A framework of learning predictions was presented in \citep{sutton2011} including out-of-policy predictions about alternative policies different from the policy the agent is behaving under in the environment\footnote{these predictions are sometimes called off-policy predictions}.
An example of an out-of-policy question is "if I keep going straight, will I drive off the road?"
These kinds of questions are very useful for decision making; if the answer is yes, then it means the road turns ahead and some corrective action is needed before the vehicle drives off the road.
A key challenge that has prevented the wide-spread adoption of predictive representations is how to learn the right predictive questions to ask \citep{awhite2015}.
Despite this challenge, the study of the brain gives us clues that the brain could not only represent knowledge in the form of sensorimotor predictions \citep{clark2013} but may even learn them with similar temporal difference learning mechanisms found in many reinforcement learning algorithms today \citep{russek2017}.

Learning out-of-policy predictions could be promising in speeding up and improving the learning of policies especially in offline learning, cf \citep{levine2020offline}.
Offline learning is about learning new policies from data that was collected in the past to solve a new task.
Offline learning requires learning out-of-policy predictions about the environment that are counter-factual "what if" questions in order to predict how the environment might change under new policies\citep{levine2020offline}.
This is a challenge that could be met by predictive representations of the form of GVFs as a framework for learning the answer to counter-factual queries referenced in \citep{levine2020offline}.
However, a model that is capable of producing all possible out-of-policy predictions is very difficult to learn; with GVFs, one could focus on learning a collection of out-of-policy predictions that are useful for learning solutions to many tasks.

Prior work in predictive representations is motivated by the predictive representation hypothesis stating that "particularly good generalization will result from representing the state of the world in terms of predictions about possible future experience" \citep{littman2002} \citep{rafols2005}.
Efforts since have focused on understanding how to learn and build an agent's representation of the world from its observations and interactions with it \citep{sutton2011}\citep{awhite2015}\citep{modayil2012}.
Two prominent directions for predictive representations exist in the literature: (1) using the predictions directly to represent the state of the agent \citep{littman2002}\citep{schaul2013}, and (2) using predictions to learn a shared representation with the agent \citep{jaderberg2018}.
In \citep{schaul2013}, general value functions (GVFs) are used as a representation of state where significant benefits are observed in partially observable grid worlds including better generalization when compared with predictive state representation (PSR) \citep{littman2002}.
Learning a shared representation through auxiliary tasks \citep{jaderberg2018} is now a very popular approach since it is very challenging to learn predictions that are sufficient to satisfy the Markov property of state.
The challenge with shared representations is that it is often not clear what kind of auxiliary tasks speed up or slow down learning; furthermore, the representations learned lack a degree of transparency in understanding how the learned representations impact control decisions.
The engineered predictive representation solution that we present provides some transparency as the predictions directly impact control decisions which is important for understanding, testing and trusting real-world applications of reinforcement learning.
Finally, only a few works have demonstrated how to learn and use predictive representations that are applied to real-world robots \citep{gunther2016} \citep{edwards2016}.

Our primary contribution focuses on investigating how predictive representations impact learning a policy with reinforcement learning and its generalization to roads and conditions not provided to the agent during training.
We focus on the challenging problem of learning to drive a vehicle between two lane markings with a camera using reinforcement learning where the state of the agent is described by a vector of GVF predictions.
The novel predictions learned are out-of-policy predictions; out-of-policy in the sense that they are fixed before learning and are different from both the optimal driving policy and the behavior policy used to learn them.
However, in many real-world applications like autonomous driving, the importance sampling ratio often used to learn out-of-policy predictions \citep{schlegel2019} cannot be computed exactly since the behavior distributions of human drivers are unknown.
Therefore, an algorithm is presented that demonstrates the efficacy in learning the predictive representations with deep learning where the importance sampling ratio is approximated since the behavior policy distribution that appears in the denominator of the importance sampling ratio is unknown.
Finally, we aim to motivate further research into the area of predictive representation learning, and its application to real-world problems such as autonomous driving.
The novel contributions of this work are summarized as follows:
\begin{itemize}
\item Demonstrate a novel predictive representation using GVFs that reveals substantial benefits in learning policies for autonomous driving in the real-world
\item Investigate how predictive representations improve learning efficiency, substantially improve the smoothness of the control decisions, and improve generalization to unseen test roads and conditions not included during training
\item Present an algorithm for learning predictive representations that approximate the importance sampling ratio for applications where the behavior policy is not known.
\end{itemize}

\section{Predictive Learning}
Let us consider a partially observable MDP described by a set of observations $O$, a set of states $S$, a set of actions $A$, transition dynamics with probability $P(s'|s,a)$ of transitioning to next state $s'$ after taking action $a$ from state $s$, and a reward $r$.
Because this setting is partially observable, the true state $s$ of the agent is hidden.
The objective of a POMDP is the classical RL objective of learning a policy $\pi$ that maximizes the future discounted sum of rewards.
A common approach to solving a POMDP is to approximate $s$ with a history of observations and actions; however, when the observations are high dimensional, learning low dimensional features with deep reinforcement learning where the policy is changing is challenging because the target is moving \citep{minh2013}.
Our objective is to achieve more stable representation learning using GVF predictions with fixed pre-defined policies which pushes the heavy burden of deep feature representation learning in RL to the easier problem of prediction learning \citep{schlegel2019}\citep{ghiassian2018}\citep{graves2019}\citep{jaderberg2018}.
For the remainder of this section, state $s$ is approximated with a history of observations and actions forming a high dimensional observation vector that contains an abundance of superfluous information.

To ask a predictive question, one must define a cumulant $c_t=c(s_t,a_t,s_{t+1})$, aka pseudo-reward, a policy $\tau(a|s)$ and continuation function $\gamma_t=\gamma(s_t,a_t,s_{t+1})$.
The answer to the predictive question is the expectation of the return $G_t$ of the cumulant $c_t$ when following policy $\tau$ defined by
\begin{equation}
V^\tau(s) = \E_{\tau}[G_t] = \E_{\tau}[\sum_{k=0}^{\infty} {(\prod_{j=0}^{k} {\gamma_{t+j+1}}) c_{t+k+1}} | s_t=s,a_t=a,a_{t+1:T-1} \sim \tau, T \sim \gamma]
\label{eq_value}
\end{equation}
where $0 \leq \gamma_t \leq 1$ \citep{sutton2011}.
The agent usually collects experience under a different behavior policy $\mu(a|s)$.
When $\mu$ is different from $\tau$, the predictive question is called an out-of-policy prediction\footnote{Some literature call this an off-policy prediction which should be distinguished from off-policy RL}.
Cumulants are commonly scaled by a factor of $1-\gamma$ when $\gamma$ is a constant in non-episodic predictions.
The GVF $V^\tau(s)$ can be approximated with any function approximator, such as a neural network, parameterized by $\theta$ to learn \eqref{eq_value}.
The parameters $\theta$ are optimized with gradient descent minimizing the following loss function
\begin{equation}
L(\theta) = \E_{s \sim d_\mu, a \sim \mu}[\rho \delta^2] = \E_{s \sim d_\mu, a \sim \tau}[\delta^2]
\label{eq_td_loss}
\end{equation}
where $\delta=\hat{v}^\tau(s;\theta) - y$ is the TD error and $\rho=\frac{\tau(a|s)}{\mu(a|s)}$ is the importance sampling ratio to correct for the difference between the policy distribution $\tau$ and behavior distribution $\mu$.
$d_\mu$ is the state distribution of the behavior policy $\mu$ and the time subscript on $c$ and $\gamma$ has been dropped to simplify notation.
Note that only the behavior policy distribution is corrected rather than the state distribution $d_\mu$.
The target $y$ is produced by bootstrapping a prediction \citep{sutton1988} of the value of the next state following policy $\tau$ given by 
\begin{equation}
y=\E_{s' \sim P}[c + \gamma \hat{v}^\tau(s';\theta)|s,a]
\label{eq_td_bootstrap}
\end{equation}
where $y$ is a bootstrapped prediction using recent parameters $\theta$ that are assumed constant in the gradient computation.
Unlike in \citep{minh2013} where target networks consisting of older parameters were needed to stabilize learning of the value function, a target network is typically not needed when learning predictions that have a fixed policy $\tau$ since learning is more stable.
The gradient of the loss function \eqref{eq_td_loss} is given by
\begin{equation}
\nabla_{\theta} L(\theta) = \E_{s \sim d_\mu, a \sim \mu}[\rho \delta \nabla_{\theta} \hat{v}^\tau(s;\theta)]
\label{eq_td_is_gradient}
\end{equation}

An alternative approach to using importance sampling ratios $\rho$ friendly to deep learning methods is to apply importance resampling \citep{schlegel2019}.
With importance resampling, a replay buffer $D$ of size $N$ is required and the gradient is estimated from a mini-batch and multiplied with the average importance sampling ratio of the samples in the buffer $\bar{\rho}=\frac{\sum_{i=1}^{N}{\rho_i}}{N}$.
The importance resampling gradient is given by
\begin{equation}
\nabla_{\theta} L(\theta) = \E_{s,a \sim D}[\bar{\rho} \delta \nabla_{\theta} \hat{v}^\tau(s;\theta)]
\label{eq_td_ir_gradient}
\end{equation}
where the transitions in the replay buffer are sampled according to $D_i=\frac{\rho_i}{\sum_{j=1}^{N}{\rho_j}}$ where $\rho_i=\frac{\tau(a_i|s_i)}{\mu(a_i|s_i)}$ for transition $i$ in the replay buffer $D$.
This approach is proven to have lower variance than \eqref{eq_td_is_gradient} with linear function approximation \citep{schlegel2019}.
An efficient data structure for the replay buffer is the SumTree used in prioritized experience replay \citep{schaul2016}.

\subsection{Estimating behavior}
A behavior policy needs to be defined to adequately explore the environment when learning GVFs.
This may be a policy defined by an expert, an evolving policy that is learned by RL, a random policy for exploring the environment, or a human driver collecting data safely.
It is common, especially in the case of human drivers, for the behavior policy distribution $\mu(a|s)$ of the agent to be unknown.
We present an algorithm to learn the behavior policy and combine it with prediction learning to achieve out-of-policy prediction learning with GVFs.
The density ratio trick is used where the ratio of two probability densities can be expressed as a ratio of discriminator class probabilities that distinguish samples from the two distributions.
Let us define a probability density function $\eta(a|s)$ for the distribution to compare to the behavior distribution $\mu(a|s)$ and class labels $y=+1$ and $y=-1$ that denote the class of the distribution that the state action pair was sampled from: $\mu(a|s)$ or $\eta(a|s)$ respectively.
A discriminator $g(a,s)$ is learned that distinguishes state action pairs from these two distributions using the cross-entropy loss.
The ratio of the densities can be computed using only the discriminator $g(a,s)$.
\begin{equation}
\begin{split}
\frac{\mu(a|s)}{\eta(a|s)} & = \frac{p(a|s,y=+1)}{p(a|s,y=-1)} = \frac{p(y=+1|a,s)p(a|s)/p(y=+1)}{p(y=-1|a,s)p(a|s)/p(y=-1)} \\
 & = \frac{p(y=+1|a,s)}{p(y=-1|a,s)} = \frac{g(a,s)}{1 - g(a,s)}
\end{split}
\label{eq_condition_density_ratio}
\end{equation}
Here we assume that $p(y=+1)=p(y=-1)$.
From this result, we can estimate $\mu(a|s)$ with $\hat{\mu}(a|s)$ as follows
\begin{equation}
\hat{\mu}(a|s) = \frac{g(a,s)}{1 - g(a,s)} \eta(a|s)
\label{eq_behavior_mu}
\end{equation}
where $\eta(a|s)$ is a known distribution over action conditioned on state.
The uniform distribution over the action is independent of state and has the advantage of being effective and easy to implement.
Alternatively, one can estimate the importance sampling ratio without defining an additional distribution $\eta$ by replacing the distribution $\eta$ with $\tau$; however, defining $\eta$ to be a uniform distribution ensured the discriminator was learned effectively across the entire action space.
The combined algorithms for training GVFs out-of-policy with an unknown behavior distribution are given in the Appendix for both the online and offline RL settings.

\section{Predictive Control for Autonomous Driving}
Once predictions have been defined and learned, our objective is to approximate the state $s$ of the agent with a low dimensional representation $\phi$ that consists of GVF predictions.
The rationale is that $\phi$ is a very compact representation and in our autonomous driving problem.
Learning a policy $\pi(a|\phi)$ could provide substantial benefits over learning $\pi(a|s)$ directly from high dimensional observations such as (1) improving learning speed of deep policies, (2) learning $\phi$ from data collected either offline or online, and (3) predicting cumulants that may be too expensive to obtain, too noisy for stable and reliable control or prohibited in certain environments.
These advantages are particularly important in autonomous driving.
Firstly, autonomous vehicles cannot be permitted to explore due to safety concerns and thus exploiting real-world data collected by both human drivers and autonomous vehicle systems for offline or batch reinforcement learning could be very valuable in learning useful policies.
Secondly, high accuracy sensors for localizing a vehicle on a map can be very expensive (e.g. DGPS) and not always available - especially in GPS-denied locations \citep{feiwu2007}.
Hence, solutions with such sensors are not easily scalable to large fleets of autonomous vehicles in diverse locations.

\begin{figure}[!h]
\centering
\includegraphics[width=8cm]{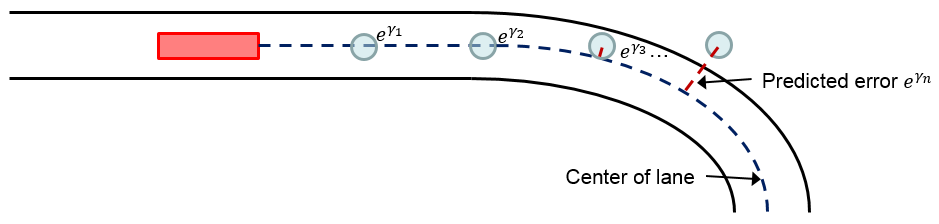}
\caption{Future predictions capturing information about the shape of the road ahead}
\label{fig_road_curvature}
\end{figure}

Several kinds of off-the-shelf methods can be applied to learning the policy $\pi(a|\phi)$: (1) online reinforcement learning with methods like deterministic policy gradient (DDPG) \citep{silver2014}, and (2) offline reinforcement learning with methods like batch-constrained Q-learning (BCQ) \citep{fujimoto2018off}.
The key insight is the way the predictive representation is crafted and learned that enables RL to learn to both steer an autonomous vehicle and modulate its speed from camera observations.
In Figure \ref{fig_road_curvature}, the center of lane is depicted; the policy $\pi$ to be learned should produce actions that cause the vehicle to follow the center of lane precisely.
Out-of-policy predictions of future lane centeredness are depicted as the deviation (or error) between the path of $\pi$ and another fixed policy $\tau$ that generates a new action that is the same (or close) to the last action taken.
This is distinctly different from prior architectures that investigate predictions where $\tau=\pi$, ie. predicting cumulants about the policy being learned \citep{barreto2017}\citep{russek2017} \citep{vanseijen2017}.
A simple thought experiment is needed to understand why predictions of future lane centeredness must be out-of-policy (or counter factual) predictions to learn a useful control policy; if we consider $\tau=\pi$ then the optimal policy $\pi$ will follow the center of line precisely resulting in all predictions being zero since there is no deviation from the lane centeredness under $\pi$ which is insufficient information to learn the optimal policy.
We postulate that out-of-policy predictions $\tau \neq \pi$ could be more important for control than on-policy predictions $\tau=\pi$ and could be what's needed to construct predictive representations that describe the state of the agent.
Out-of-policy predictions can be thought of as anticipated future "errors" that allow controllers to take corrective actions before the errors occur.

\begin{figure}[!h]
	\centering
	\begin{subfigure}[b]{4cm}
		\includegraphics[width=4cm]{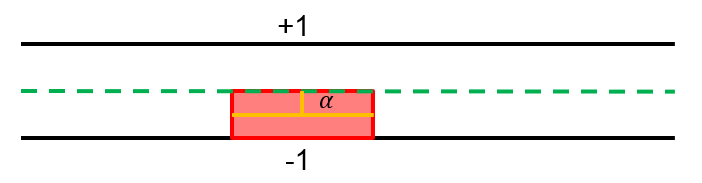}
		\caption{Lane centeredness}
		\label{fig_trackpos}
	\end{subfigure}
	\begin{subfigure}[b]{4cm}
		\includegraphics[width=4cm]{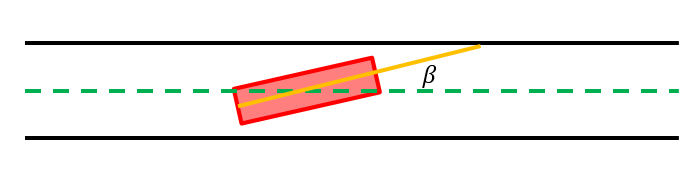}
		\caption{Road angle}
		\label{fig_roadangle}
	\end{subfigure}
\caption{(a) the lane centeredness position $\alpha$ is the distance from the center of the lane to the center of the vehicle. (b) the road angle $\beta$ is the angle between the direction of the vehicle and the direction of the road.}
\label{fig_road_predictions}
\end{figure}

To drive an autonomous vehicle, we learn predictions of lane centeredness and road angle using available (and potentially noisy) localization techniques and a map.
Cumulants for lane centeredness  $\alpha$ and road angle $\beta$ are depicted in Figure \ref{fig_road_predictions}.
Multiple time scales are learned, as depicted in Figure \ref{fig_road_curvature}, which allow the agent to predict how the road curves in the future in a very compact representation from just images.
$\phi(s)$ is thus the concatenation of lane centeredness predictions over multiple time scales, road angle predictions over multiple time scales, current speed of the vehicle, and the last action taken.
It is worth pointing out that $\phi(s)$ consists of both predictions and two pieces of low dimensional information that are difficult to ascertain from predictions of lane centeredness and road angle; in particular the last action is crucial because the policy $\tau$ depends on the last action.

\section{Experiments}
The policies using predictive representations that control the steering angle and vehicle speed based on a front facing camera are learned and evaluated in two environments: the TORCS racing simulator \citep{wymann2013} and a real-world Jackal robot.
Our objective is to understand the impact predictive representations have on (1) learning, (2) generalization and robustness, and (3) smoothness of control.
The policies that are learned in TORCS are evaluated on unseen test road periodically during simultaneous training of predictions and RL driving policy.
For experiments on the Jackal robot, both representations and policies are trained offline from exploration data collected in the real world; the policy was trained with BCQ \citep{fujimoto2018off}.
The Jackal robot was then evaluated on test roads different and more complex from the training data including damaged and distracted roads to test the robustness of the policies in the real-world.

\subsection{TORCS Simulator}
The proposed solution has three variants: (1) myopic predictions called GVF-0.0-DDPG where $\gamma=0$ which is similar to the method in \citep{chen2015}, (2) future predictions called GVF-0.95-DDPG where $\gamma=0.95$, and (3) multiple time scale predictions called GVF-DDPG with $\gamma=[0.0, 0.5, 0.9, 0.95, 0.97]$.
For details on how the predictive representations are learned, refer to the Appendix.
We compare with a kinematic-based steering approach \citep{paden2016} and two variants of DDPG.
The kinematic-based steering approach is a classical controller that using lane centeredness $\alpha$ and road angle $\beta$ as controlled process variables where steering and speed are controlled independently.
The DDPG variants are (1) DDPG-Image where DDPG observations are the same as the GVF-DDPG method, and (2) DDPG-LowDim where DDPG appends lane centeredness $\alpha$ and road angle $\beta$ to the observation vector.
All learned policies output steering angle and vehicle speed commands and observe a history of two images, the current vehicle speed and the last action. 

\begin{figure}[!h]
	\centering
	\includegraphics[width=14cm]{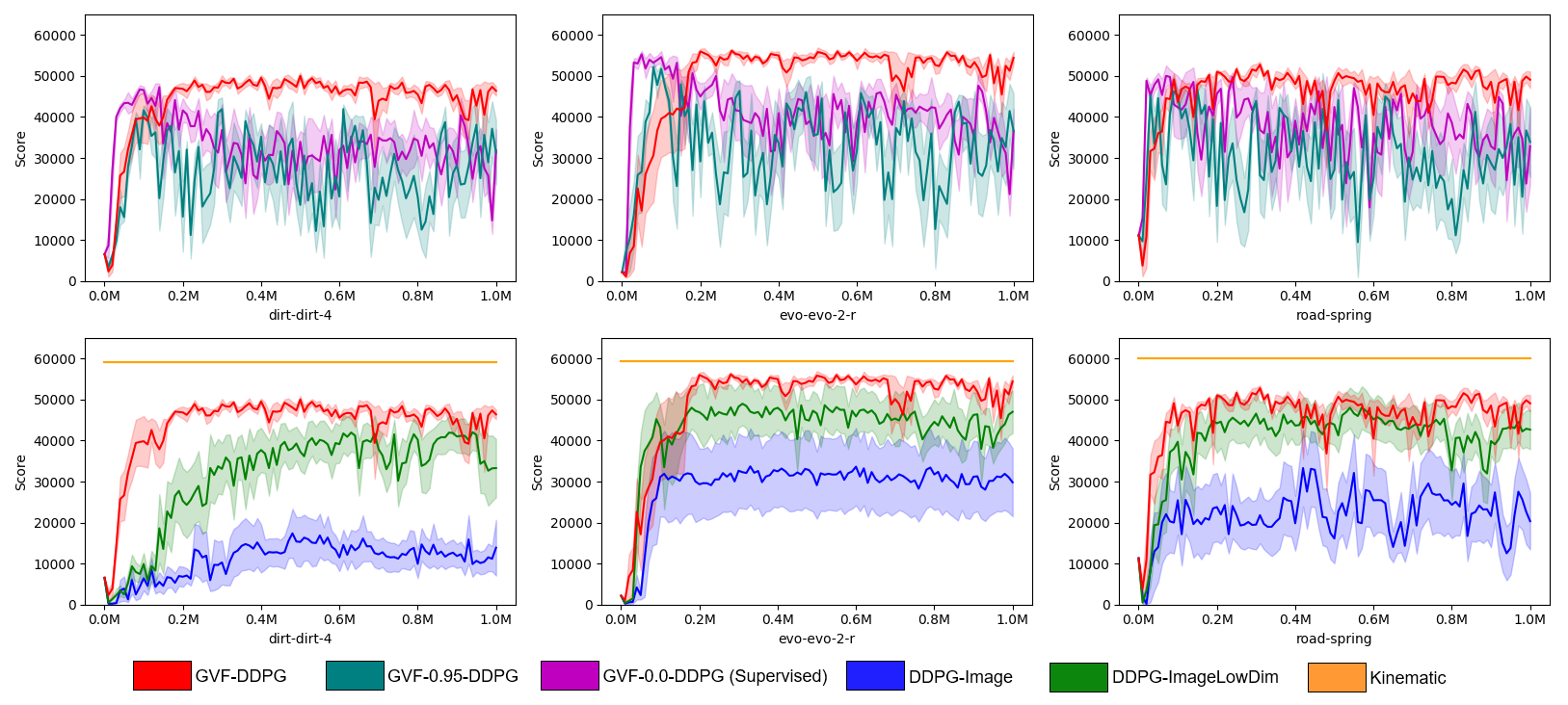}
	\caption{Test scores (accumulated reward) evaluated every 1000 steps during training for dirt-dirt-4, evo-evo-2 and road-spring}
	\label{fig_summary_scores}
\end{figure}

\begin{figure}[!h]
	\centering
	\includegraphics[width=14cm]{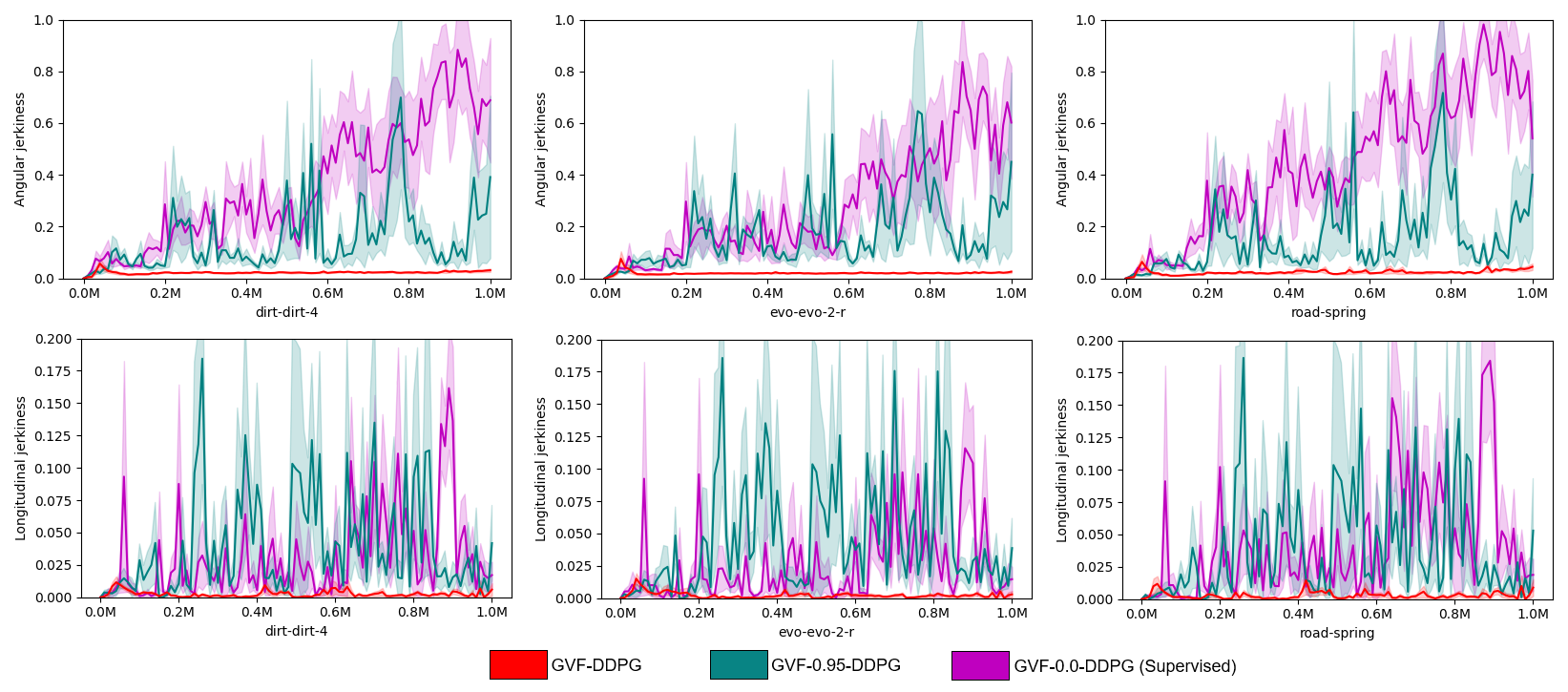}
	\caption{Angular and longitudinal jerkiness evaluated every 1000 steps during training for dirt-dirt-4, evo-evo-2 and road-spring}
	\label{fig_jerkiness}
\end{figure}

The agents were trained on 85\% of 40 tracks available in TORCS.
The rest of the tracks were used for testing (6 in total) to measure the generalization performance of the policies.
Results are repeated over 5 runs for each method.
Only three of the tracks were successfully completed by at least one learned agent and those are reported here; refer to the Appendix for more detailed results.
The reward in the TORCS environment is given by $r_t=0.0002 v_t (\cos{\beta_t} + |\alpha_t|)$ where $v_t$ is the speed of the vehicle in km/h, $\beta_t$ is the angle between the road direction and the vehicle direction, and $\alpha_t$ is the current lane centeredness.
The policies were evaluated on test roads at regular intervals during training as shown in Figures \ref{fig_summary_scores} and \ref{fig_jerkiness}.

The GVF-DDPG agent achieved the best solution of all the learned agents and approached the performance of the baseline. 
The GVF-0.0-DDPG and GVF-0.95-DDPG variations initially learned very good solutions but then diverged indicating that one prediction may not be enough to control both steering angle and vehicle speed.
This agrees with findings in psychological studies of how humans drive a vehicle \citep{salvucci2004}.
Despite an unfair advantage provided by DDPG-LowDim with the inclusion of lane centeredness and road angle in the observation vector, GVF-DDPG still outperforms both variants of DDPG on many of the test roads.
DDPG-Image was challenging to tune and train due to instability in learning that is widely known in RL literature.
The angular jerkiness and longitudinal jerkiness are defined as the standard deviation in the change in steering action and speed action respectively.
Only GVF-DDPG with multiple time scale predictions is able to achieve extraordinarily smooth control that is only matched by the classical control baseline; both DDPG-Image and DDPG-LowDim fail to provide smooth control policies.
Additional results are found in the Appendix.

\subsection{Jackal Robot}
We then applied the same methods to a Jackal robot using only data collected in the real-world.
The proposed solution, simply called GVF, is a combination of learning a predictive representation with GVFs and offline reinforcement learning with BCQ \citep{fujimoto2018off}.
Two baselines are designed: (1) a classical controller using existing model predictive control (MPC) software available for the robot in ROS packages that uses a map and laser scanner for localization, and (2) a offline deep reinforcement learning solution called E2E learned end-to-end with BCQ directly from images.
Offline reinforcement learning was convenient since it minimized wear and tear on the robot over long periods of interaction, reduced safety concerns due to exploration and reduced the need of human supervision during training since the robot's battery needed to be charged very 4 hours.
All of these considerations made online learning with RL impractical.
The GVF and E2E methods were trained with the same data and observations for a total of 10M updates.
For GVF, the first half of the updates were dedicated to learning the predictive representation and the second half of the updates were dedicated to learning the policy with BCQ.
The observations of the learned policies consisted of a history of 2 images, current vehicle speed, and last action.
The learned polices output the steering angle and desired speed.
The reward is $r_t=v_t (\cos{\beta_t} + |\alpha_t|)$ where $v_t$ is the speed of the vehicle in km/h, $\beta_t$ is the angle between the road direction and the vehicle direction, and $\alpha_t$ is the current lane centeredness.

The training data consisted of 6 training roads and 3 test roads where each of the 3 test roads had damaged variants.
The test roads were (1) a rectangle-shaped road with rounded outer corners, (2) an oval-shaped road, and (3) a road with multiple turns with complexity not observed by the agent during training.
Refer to the Appendix for more details in the experimental setup and training.

Highlights of the most interesting results are provided in Tables 
\ref{table_summary_jackal_result_gvf_vs_e2e}, \ref{table_summary_jackal_result_gvf_vs_mpc} and \ref{table_summary_jackal_result_N_vs_D}.
GVF was compared to E2E in Table \ref{table_summary_jackal_result_gvf_vs_e2e} where GVF was found to be far superior to E2E; the E2E policy had an average speed roughly half of that of the GVF and failed to generalize.
E2E was able to drive almost an entire loop of the rectangle road in the counter clockwise (CCW) direction but went out of the lane and was unable to recover; moreover, in the clockwise (CW) direction, E2E did not generalize in the reverse direction and failed immediately.
In Table \ref{table_summary_jackal_result_gvf_vs_mpc}, GVF was compared to MPC where GVF was found to produce superior performance in nearly all metrics at a high target speed of 0.4 m/s.
MPC was difficult to tune for 0.4 m/s; At 0.25 m/s, the gap in performance between GVF and MPC was lessened.
Table \ref{table_summary_jackal_result_N_vs_D} highlights the similarity in performance when the track is damaged demonstrating how well the GVF method generalizes.
Refer to the Appendix for more detailed results and plots.

\begin{table}
\centering
\caption{Comparison of GVF and E2E on Rectangle test road with 0.4 m/s target speed}
\label{table_summary_jackal_result_gvf_vs_e2e}
\begin{tabular}{|l|l|l|l|l|l|l|l|} 
\hline
\multicolumn{1}{|c|}{\begin{tabular}[c]{@{}c@{}}Method\end{tabular}} & \multicolumn{1}{c|}{\begin{tabular}[c]{@{}c@{}}Direction\end{tabular}} & \multicolumn{1}{c|}{\begin{tabular}[c]{@{}c@{}}Reward\\/s \textuparrow \end{tabular}} & \multicolumn{1}{c|}{\begin{tabular}[c]{@{}c@{}}Average\\ speed \textuparrow \end{tabular}} & \multicolumn{1}{c|}{\begin{tabular}[c]{@{}c@{}}Offcentered\\ -ness \textdownarrow~\end{tabular}} & \multicolumn{1}{c|}{\begin{tabular}[c]{@{}c@{}}Absolute\\ road angle \textdownarrow \end{tabular}} & \multicolumn{1}{c|}{\begin{tabular}[c]{@{}c@{}}Near out\\ of lane \textdownarrow \end{tabular}}  \\ 
\hline
GVF & CCW & \textbf{2.6835 } & \textbf{0.3205 } & \textbf{0.1345 } & \textbf{0.1315 } & \textbf{0.0 } \\
E2E & CCW & 1.2578 & 0.1816 & 0.2558 & 0.2414 & 0.0376 \\ 
\hline
GVF & CW & \textbf{2.2915} & \textbf{0.3140} & \textbf{0.2217} & \textbf{0.1586} & \textbf{0.0} \\
E2E & CW & -0.1302 & 0.1710 & 0.9927 & 0.3034 & 0.5418 \\
\hline
\end{tabular}
\end{table}

\begin{table}
\centering
\caption{Comparison of GVF and MPC in CCW direction with 0.4 m/s target speed}
\label{table_summary_jackal_result_gvf_vs_mpc}
\begin{tabular}{|l|l|l|l|l|l|l|l|} 
\hline
 & \multicolumn{1}{|c|}{\begin{tabular}[c]{@{}c@{}}Method \end{tabular}} & \multicolumn{1}{|c|}{\begin{tabular}[c]{@{}c@{}}Reward \\ /s \textuparrow \end{tabular}} & \multicolumn{1}{c|}{\begin{tabular}[c]{@{}c@{}}Offcenter-\\edness \textdownarrow\end{tabular}} & \multicolumn{1}{c|}{\begin{tabular}[c]{@{}c@{}}Absolute\\ road angle \textdownarrow \end{tabular}} & \multicolumn{1}{c|}{\begin{tabular}[c]{@{}c@{}}Near out\\ of lane \textdownarrow \end{tabular}} & \multicolumn{1}{c|}{\begin{tabular}[c]{@{}c@{}}Speed\\ comfort \textuparrow \end{tabular}} & \multicolumn{1}{c|}{\begin{tabular}[c]{@{}c@{}}Steering\\ comfort \textuparrow \end{tabular}} \\
\hline
\multirow{2}{*}{\begin{tabular}[c]{@{}l@{}}Rect-\\angle \end{tabular}}
 & GVF & \textbf{2.6835} & \textbf{0.1345} & \textbf{0.1315} & \textbf{0.0} & \textbf{-0.0356} & \textbf{-0.2251} \\
 & MPC & 0.9700 & 0.5252 & 0.1943 & 0.2042 & -0.0832 & -1.2542 \\ 
\hline
\multirow{2}{*}{\begin{tabular}[c]{@{}l@{}}Oval \end{tabular}}
 & GVF & \textbf{2.4046} & \textbf{0.2754} & 0.2125 & \textbf{0.0145} & \textbf{-0.0348} & \textbf{-0.2191} \\
 & MPC & 0.8928 & 0.5293 & \textbf{0.1963} & 0.227 & -0.1026 & -1.4119 \\ 
\hline
\multirow{2}{*}{\begin{tabular}[c]{@{}l@{}}Comp-\\lex \end{tabular}}
 & GVF & \textbf{2.3501} & \textbf{0.2221} & \textbf{0.1817} & \textbf{0.0} & \textbf{-0.0341} & \textbf{-0.2272} \\
 & MPC & 0.7172  & 0.6407 & 0.2131 & 0.3894 & -0.0625 & -1.2133 \\ 
\hline
\end{tabular}
\end{table}

\begin{table}
\centering
\caption{Effect of damaged lanes on GVF performance in CCW direction with 0.4 m/s target speed}
\label{table_summary_jackal_result_N_vs_D}
\begin{tabular}{|l|l|l|l|l|l|l|l|} 
\hline
 & \multicolumn{1}{|c|}{\begin{tabular}[c]{@{}c@{}}Damage \end{tabular}} & \multicolumn{1}{c|}{\begin{tabular}[c]{@{}c@{}}Reward\\/s \textuparrow \end{tabular}} & \multicolumn{1}{c|}{\begin{tabular}[c]{@{}c@{}}Offcenter-\\edness \textdownarrow \end{tabular}} & \multicolumn{1}{c|}{\begin{tabular}[c]{@{}c@{}}Absolute\\ road angle \textdownarrow \end{tabular}} & \multicolumn{1}{c|}{\begin{tabular}[c]{@{}c@{}}Near out\\ of lane \textdownarrow \end{tabular}} & \multicolumn{1}{c|}{\begin{tabular}[c]{@{}c@{}}Speed\\ comfort \textuparrow \end{tabular}} & \multicolumn{1}{c|}{\begin{tabular}[c]{@{}c@{}}Steering\\ comfort \textuparrow \end{tabular}}  \\ 
\hline
\multirow{2}{*}{\begin{tabular}[c]{@{}l@{}}Recta-\\ngle\end{tabular}}
 & No & 2.6835 & \textbf{0.1345 } & \textbf{0.1315 } & 0.0 & \textbf{-0.0356 } & \textbf{-0.2251 } \\
 & Yes & \textbf{2.7407 } & 0.1358 & 0.1351 & 0.0 & -0.0383 & -0.2303 \\ 
\hline
\multirow{2}{*}{Oval}
 & No & \textbf{2.4046 } & \textbf{0.2754 } & 0.2125 & \textbf{0.0145 } & -0.0348 & -0.2191 \\
 & Yes & 2.0728 & 0.3285 & \textbf{0.2089 } & 0.0719 & \textbf{-0.0334 } & \textbf{-0.2094 } \\ 
\hline
\multirow{2}{*}{\begin{tabular}[c]{@{}l@{}} Comp-\\lex\end{tabular}}
 & No & \textbf{2.3501 } & \textbf{0.2221 } & \textbf{0.1817 } & \textbf{0.0 } & \textbf{-0.0341 } & \textbf{-0.2272 } \\
 & Yes & 2.1059 & 0.3125 & 0.2365 & 0.0942 & -0.0437 & -0.2897 \\
\hline
\end{tabular}
\end{table}

\section{Conclusions}
An investigation into the problem of driving a vehicle between lane markings reveals that a small collection of out-of-policy predictions of the future at different time scales provides many benefits in learning general policies with reinforcement learning with applications to real-world robots and autonomous driving.
By testing our policies on roads not shown to the agent during training, including damaged roads and roads with distracted markings, we find that our novel predictive representation improves learning speed, generalization in the real-world, and much smoother control decisions.
In addition, it was found that learning a predictive representation of future lane centeredness and road angle over multiple time scales was required to achieve better performance in agreement with the two-point model described in \citep{salvucci2004} based on psychological studies performed decades ago on understanding how humans drive.

Learning to drive from images end-to-end is not new \citep{bojarski2016}\citep{garimella2017}\citep{chen2017}\citep{sallab2017}\citep{chi2017}.
However, a serious challenge has been ensuring the robustness of the controller and its generalization to roads and conditions not seen in the training data \citep{bojarski2016}.
In this work, we show that policies that generalize can be learned from data collected offline where the learning of the predictive representation can be decoupled from the learning of the policy.
Our real-world experiments demonstrated significant improvements in learning good policies with  offline reinforcement learning methods like BCQ \citep{fujimoto2018off}.
We conclude that predictive representations can be used to represent the agent's knowledge of the world, supporting claims in \citep{awhite2015} and \citep{sutton2011}, where our contributions are in presenting ways to learn and exploit that predictive knowledge in autonomous driving.
The improvements offered in our algorithm of learning predictive representations from data where the behavior policy is unknown has applications for applying reinforcement learning to autonomous driving in the real-world, especially as large volumes of data could potentially be captured by human drivers of all skill levels at scale that include valuable rare events like collisions.
Challenging problems however include evaluating predictive representations and learning the predictive questions.
Future work includes building a general framework to learn the predictive questions themselves and investigate how shared representations offered by auxiliary tasks compare to direct representations.
One obviously advantage of direct representations is that the representation is significantly smaller (8 predictions + 2) compared with potentially hundreds of features that are typically learned for shared representations.

\clearpage
\bibliographystyle{plainnat}
\bibliography{references}

\clearpage
\section{Appendix}

\subsection{Predictive Learning Algorithms}

The algorithm used for learning GVF predictions online through interaction with an environment is given in Algorithm \ref{alg_gvf_offpolicy_train_without_mu_online}.

\begin{algorithm}
\caption{Online Out-of-policy GVF training algorithm with unknown $\mu(a|s)$}
\label{alg_gvf_offpolicy_train_without_mu_online}
\begin{algorithmic}[1]
\State Initialize $\hat{v}^\tau$, $g(a,s)$, $\eta(a|s)$, and replay memory $D$
\State Observe initial state $s_0$
\For{$t=0$,$T$}
  \State Sample action $a_t$ from unknown $\mu(a_t|s_t)$
  \State Execute action $a_t$ and observe state $s_{t+1}$
  \State Compute cumulant $c_{t+1}=c(s_t,a_t,s_{t+1})$
  \State Compute continuation $\gamma_{t+1}=\gamma(s_t,a_t,s_{t+1})$
  \State Estimate behavior density value $\hat{\mu}(a_t|s_t)=\frac{g(a_t,s_t)}{1-g(a_t,s_t)}\eta(a_t|s_t)$
  \State Estimate importance sampling ratio $\rho_t=\frac{\tau(a_t|s_t)}{\hat{\mu}(a_t|s_t)}$
  \State Store transition $(s_t,a_t,c_{t+1},\gamma_{t+1},s_{t+1},\rho_t)$ in $D$
  \State Compute average importance sampling ratio in replay buffer $D$ of size $n$ with $\bar{\rho}=\frac{1}{n}\sum_{j=1}^{n}{\rho_j}$
  \State Sample random minibatch $A$ of transitions $(s_i,a_i,c_{i+1},\gamma_{i+1},s_{i+1})$ from $D$ according to probability $\frac{\rho_i}{\sum_{j=1}^{n}\rho_j}$
  \State Compute $y_i = c_{i+1} + \gamma_{i+1} \hat{v}^\tau(s_{i+1};\theta)$ for minibatch $A$
  \State Update parameters $\theta$ using gradient descent on \eqref{eq_td_loss} with gradient \eqref{eq_td_ir_gradient} over the minibatch $A$
  \State Sample random minibatch $B$ of state action pairs $(s_i,a_i)$ from $D$ according to a uniform probability and assign label $z=1$ to each pair
  \State Randomly select half the samples in the minibatch $B$ replacing the action with $a_t \sim \eta(a|s)$ and label with $z=0$ and storing the updated samples in $\hat{B}$
  \State Update behavior discriminator $g(a,s)$ with labels $z$ in the modified minibatch $\bar{B}$ using binary cross-entropy loss
\EndFor
\end{algorithmic}
\end{algorithm}

With minor modifications, an offline version of the algorithm can be derived.
This algorithm learns by reading the data in sequence and populating a replay buffer just as it would in online learning; the only difference is that the offline algorithm returns the action taken in the data.
This allows the same algorithm and code for learning GVF predictions to be used in either online or offline learning settings.

\begin{algorithm}
\caption{Offline Out-of-policy GVF training algorithm with unknown $\mu(a|s)$}
\label{alg_gvf_offpolicy_train_without_mu_offline}
\begin{algorithmic}[1]
\State Initialize $\hat{v}^\tau$, $g(a,s)$, $\eta(a|s)$, and replay memory $D$,
\State Obtain the first state in the data file $s_0$
\For{$t=0$,$T$}
  \State Obtain action $a_t$ recorded in the data file that sampled from an unknown $\mu(a_t|s_t)$
  \State Obtain next state $s_{t+1}$ from the data file
  \State Compute cumulant $c_{t+1}=c(s_t,a_t,s_{t+1})$
  \State Compute continuation $\gamma_{t+1}=\gamma(s_t,a_t,s_{t+1})$
  \State Estimate behavior density value $\hat{\mu}(a_t|s_t)=\frac{g(a_t,s_t)}{1-g(a_t,s_t)}\eta(a_t|s_t)$
  \State Estimate importance sampling ratio $\rho_t=\frac{\tau(a_t|s_t)}{\hat{\mu}(a_t|s_t)}$
  \State Store transition $(s_t,a_t,c_{t+1},\gamma_{t+1},s_{t+1},\rho_t)$ in $D$
  \State Compute average importance sampling ratio in replay buffer $D$ of size $n$ with $\bar{\rho}=\frac{1}{n}\sum_{j=1}^{n}{\rho_j}$
  \State Sample random minibatch $A$ of transitions $(s_i,a_i,c_{i+1},\gamma_{i+1},s_{i+1})$ from $D$ according to probability $\frac{\rho_i}{\sum_{j=1}^{n}\rho_j}$
  \State Compute $y_i = c_{i+1} + \gamma_{i+1} \hat{v}^\tau(s_{i+1};\theta)$ for minibatch $A$
  \State Update parameters $\theta$ using gradient descent on \eqref{eq_td_loss} with gradient \eqref{eq_td_ir_gradient} over the minibatch $A$
  \State Sample random minibatch $B$ of state action pairs $(s_i,a_i)$ from $D$ according to a uniform probability and assign label $z=1$ to each pair
  \State Randomly select half the samples in the minibatch $B$ replacing the action with $a_t \sim \eta(a|s)$ and label with $z=0$ and storing the updated samples in $\hat{B}$
  \State Update behavior discriminator $g(a,s)$ with labels $z$ in the modified minibatch $\bar{B}$ using binary cross-entropy loss
\EndFor
\end{algorithmic}
\end{algorithm}

\section{TORCS Experiments}
TORCS is a racing simulator used for learning to drive.
All opponent vehicles were removed for these experiments as well as roads that were sloped.
The goal of the agent is to maximize the future accumulation of the following reward: $r_t=0.0002 v_t (\cos{\beta_t} + |\alpha_t|)$ where $v_t$ is the speed of the vehicle in km/h, $\beta_t$ is the angle between the road direction and the vehicle direction, and $\alpha_t$ is the current lane centeredness.
Termination occurs when either the agent leaves the lane or the maximum number of steps has been reached (1200 steps = 120 seconds) triggering a reset of the environment.
Upon reset, a priority sampling method is used during training to select the next road to train on.
The probability of sampling road $i$ during a reset is given by 

\begin{equation}
\frac{e^{-\frac{n_i}{\kappa}}}{\sum_{j=1}^{N}{e^{-\frac{n_j}{\kappa}}}}
\label{eq_track_sample}
\end{equation}

where $n_i$ is the number of steps that the agent was able to achieve the last time the road was sampled and $\kappa$ controls the spread of the distribution.
A value of $\kappa=\frac{1}{N}\sum_{j=1}^{N}{n_j}$ was found to perform well.
The initial probabilities are equal for all roads.
This improved the efficiency in learning for all learned methods.

The TORCS environment was modified to provide higher resolution images in grayscale rather than RGB with most of the image above the horizon cropped out of the image.
The grayscale images were 128 pixels wide by 64 pixels high.
This allowed the agent to see more detail farther away which is very helpful in making long term predictions and is beneficial to both policy gradient methods and predictive learning.

\subsection{Training}
Two main learning algorithms are compared, along with their variants:  our proposed GVF-DDPG (general value function deep deterministic policy gradient) and DDPG.
The parameters used to train the methods will be described in more detail here.

\subsubsection{GVF-DDPG Training}
The predictive learning approach presented in this paper is called GVF-DDPG (general value function deep deterministic policy gradient).
Exploration followed the same approach as \citep{lillicrap2016} where an Ornstein Uhlenbeck process \citep{uhlenbeck1930} was used to explore the road; the parameters of the process ($\theta=1.0$, $\sigma=0.1$, $dt=0.01$) were tuned to provide a gradual wandering behavior on the road without excessive oscillations in the action.
This improved the learning of the off-policy predictions for GVF-DDPG since the behavior policy $\mu(a|s)$ was closer to the target policy $\tau(a|s)$ of the predictions.

The GVF-DDPG approach learned 8 predictions:  4 predictions of lane centeredness $\alpha$, and 4 predictions of road angle $\beta$.
Each of the 5 predictions had different values of $\gamma$ for different temporal horizons:  0.0, 0.5, 0.9, 0.95, 0.97.
This allowed the agent to predict how the road will turn in the future providing the necessary look ahead information for the agent to control effectively.
The GVF predictors shared the same deep convolutional neural network where the convolutional layers were identical to the architecture in \citep{bojarski2016} followed by three fully connected layers of 512, 384 and 8 outputs, respectively.
The behavior estimator $\mu(a|s)$ also used a very similar neural network with identical convolutional layers followed by three fully connected layers of 512, 256, and 1 output, respectively.
The GVF predictors and behavior estimator were both given the current image, previous image, current speed and the last action taken.

The policy network of the DDPG agent was a small neural network with two branches for each action -- steering angle and target speed -- where each branch received the 8 predictions and last action as input that fed into three fully connected layers of 256, 128, and 1 output.
ReLU activation was used for the hidden layers while linear activation was used on the output layer followed by a clip to the range $[-1.0, 1.0]$.
The output target speed passed through a linear transformation to change the range of values from $[-1.0, 1.0]$ to $[0.5, 1.0]$.
The structure of the action-value network was the same as the policy network, except it received the 8 predictions, last action, and current action as input as well as used a linear activation on the output layer.
Additionally, the steering and target speed input into both policy and action-value networks were normalized to range $[-1.0, 1.0]$. 

A replay buffer of size 100000 was used with a warmup of 10000 samples.
In order to not bias the replay buffer, the last layers in each branch of the policy network were initialized with a uniform distribution $[-1e^{-3}, 1e^{-3}]$ for the weight and 0 for the bias.
The learning rates for the policy network, action-value network, predictors and behavior policy network were $1e^{-6}$, $1e^{-4}$, $1e^{-4}$, and $1e^{-4}$ respectively.
Target networks \citep{lillicrap2016} were used for the action-value and policy networks with $\tau=0.001$ in order to make the bootstrapped prediction of the action-values more stable.
The reward for was scaled by 0.0002 to scale the action-values to fall within the range $[-1.0,1.0]$.

\subsubsection{Baseline DDPG Training}
The two DDPG (deep deterministic policy gradient) \citep{lillicrap2016} baselines were trained nearly identically where the only difference was the information provided in the observation.
The first method called DDPG-Image is a vision-based approach where the image, current speed, and last action are provided to the agent; the only information available to the agent about the road is supplied via images.
The second agent called DDPG-LowDim added lane centeredness $\alpha$ and road angle $\beta$ to the observation used by DDPG-Image.
It should be noted that DDG-LowDim was the only learned approach that had access to $\alpha$ and $\beta$ during testing, whereas GVF-DDPG and DDPG-Image only need the information to compute the cumulant and reward during training.

Training setup for DDPG was largely the same as GVF-DDPG.
An Ornstein Uhlenbeck process \citep{uhlenbeck1930} was used to explore the road ($\theta=1.0$, $\sigma=0.1$, and $dt=0.01$).
Experimentally, it was found that the exploration parameters did not affect the learning performance of DDPG that much.
Target networks were used with $\tau=0.001$.
The learning rates of the action-value and policy networks were the same as those of GVF-DDPG.
The architecture of the behavior network for GVF-DDPG was used for action-value and policy network of DDPG, except that two fully connected branches with layer sizes of 512, 256 and 1 output were used for each action of the policy network.
As with GVF-DDPG, linear activation followed by a clip to the range $[-1.0, 1.0]$ was used for each output of the policy network and ReLU activation was for all other layers.
The linear transformation of the target speed and normalization of the input action was accomplished in the same way as GVF-DDPG.

The low dimensional state information was fed through a separate branch of 12 neurons that were then merged with the input into the fully connected layer of 512 neurons.
This method of merging presents challenges due to the potential mismatch of the statistical properties of the branches but that did not seem to present a significant challenge in learning.
Future work would be to find better ways to bridge these two different pieces of information.

\subsection{Experimental Results}

\begin{figure}[!h]
	\centering
	\begin{subfigure}[b]{10cm}
		\includegraphics[width=10cm]{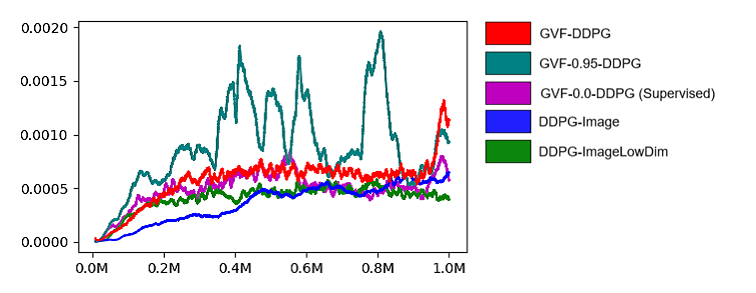}
		\caption{Action-Value}
	\end{subfigure}
	\\
	\begin{subfigure}[b]{5cm}
		\includegraphics[width=5cm]{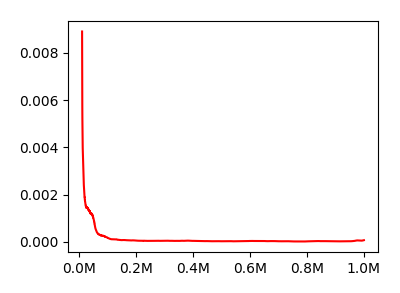}
		\caption{Predictors}
	\end{subfigure}
	\begin{subfigure}[b]{5cm}
		\includegraphics[width=5cm]{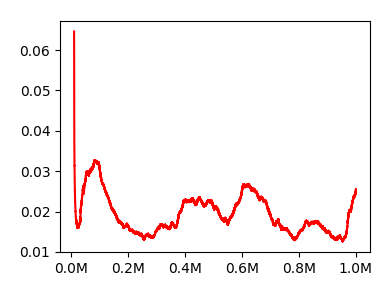}
		\caption{Behavior}
	\end{subfigure}
	\caption{Learning curves for (a) Q-values of the DDPG agents, (b) mean squared TD (temporal difference) errors of the GVF predictors, and (c) MSE of the behavior model estimator}
	\label{fig_learning_curve}
\end{figure}

\begin{figure}[!h]
	\centering
	\includegraphics[width=12cm]{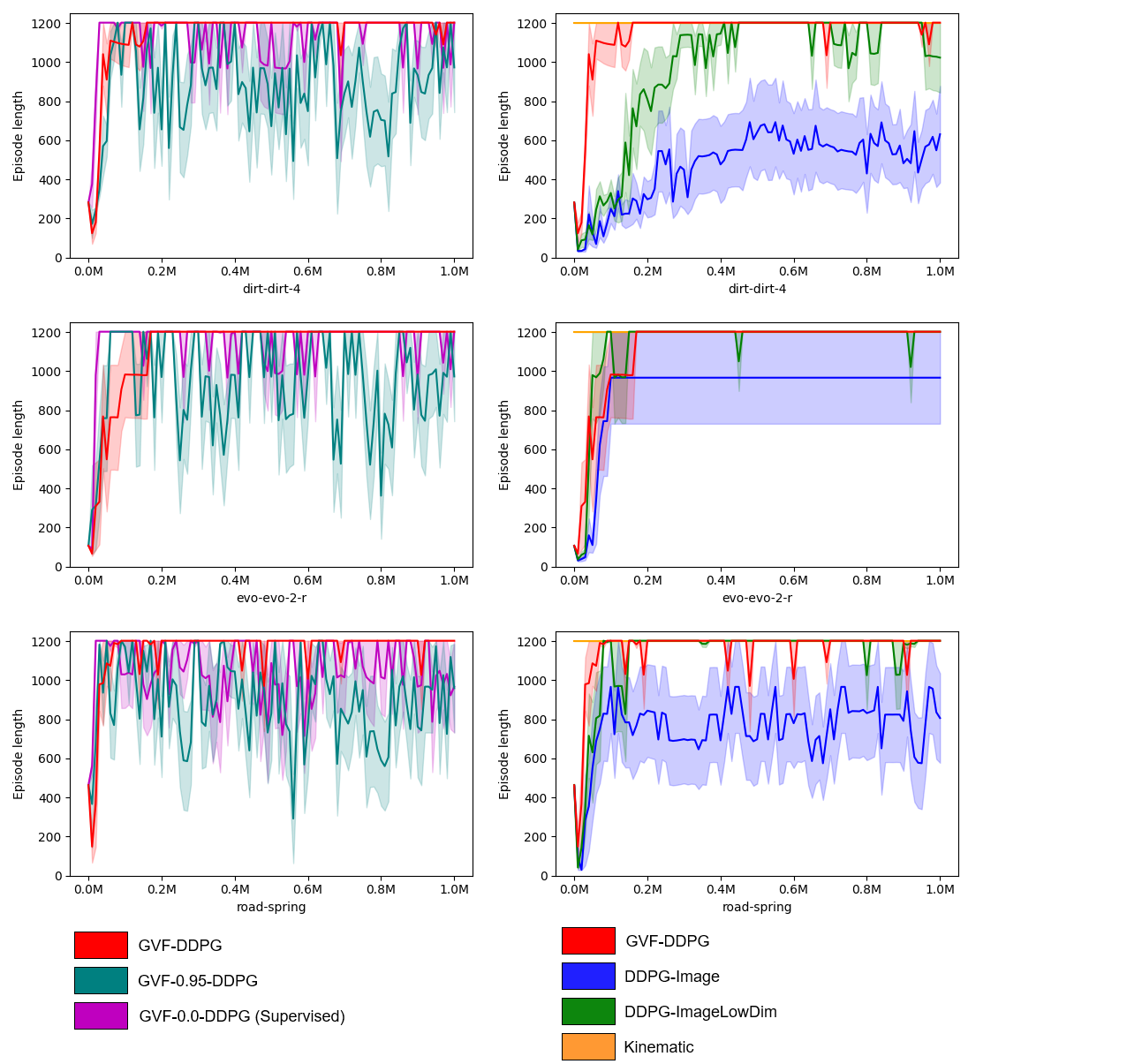}
	\caption{Mean episode length during training for dirt-dirt-4, evo-evo-2 and road-spring}
	\label{fig_episode_length}
\end{figure}

\begin{figure}[!h]
	\centering
	\includegraphics[width=12cm]{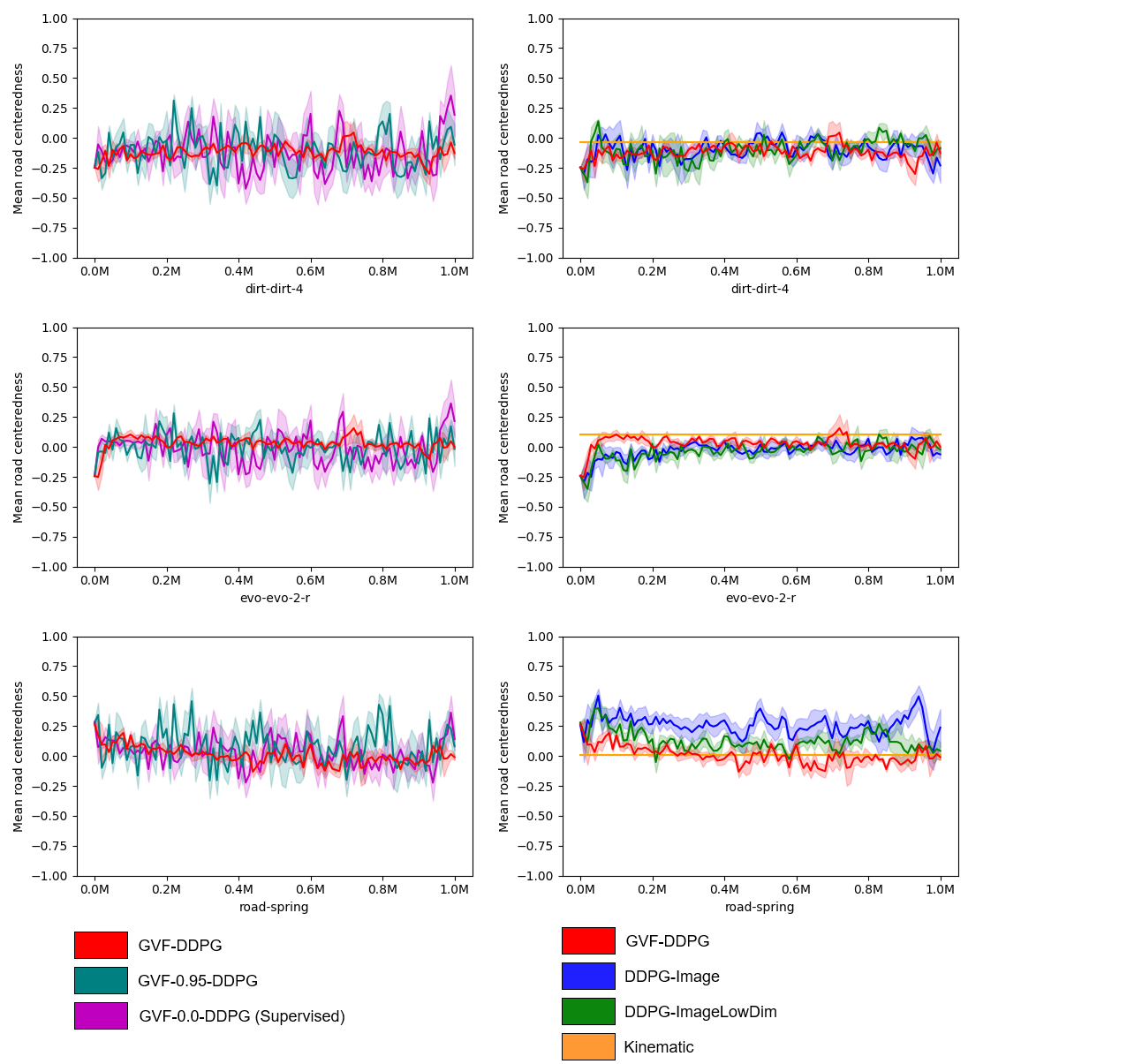}
	\caption{Mean lane centeredness during training for dirt-dirt-4, evo-evo-2 and road-spring}
	\label{fig_road_centeredness}
\end{figure}

\begin{figure}[!h]
	\centering
	\includegraphics[width=12cm]{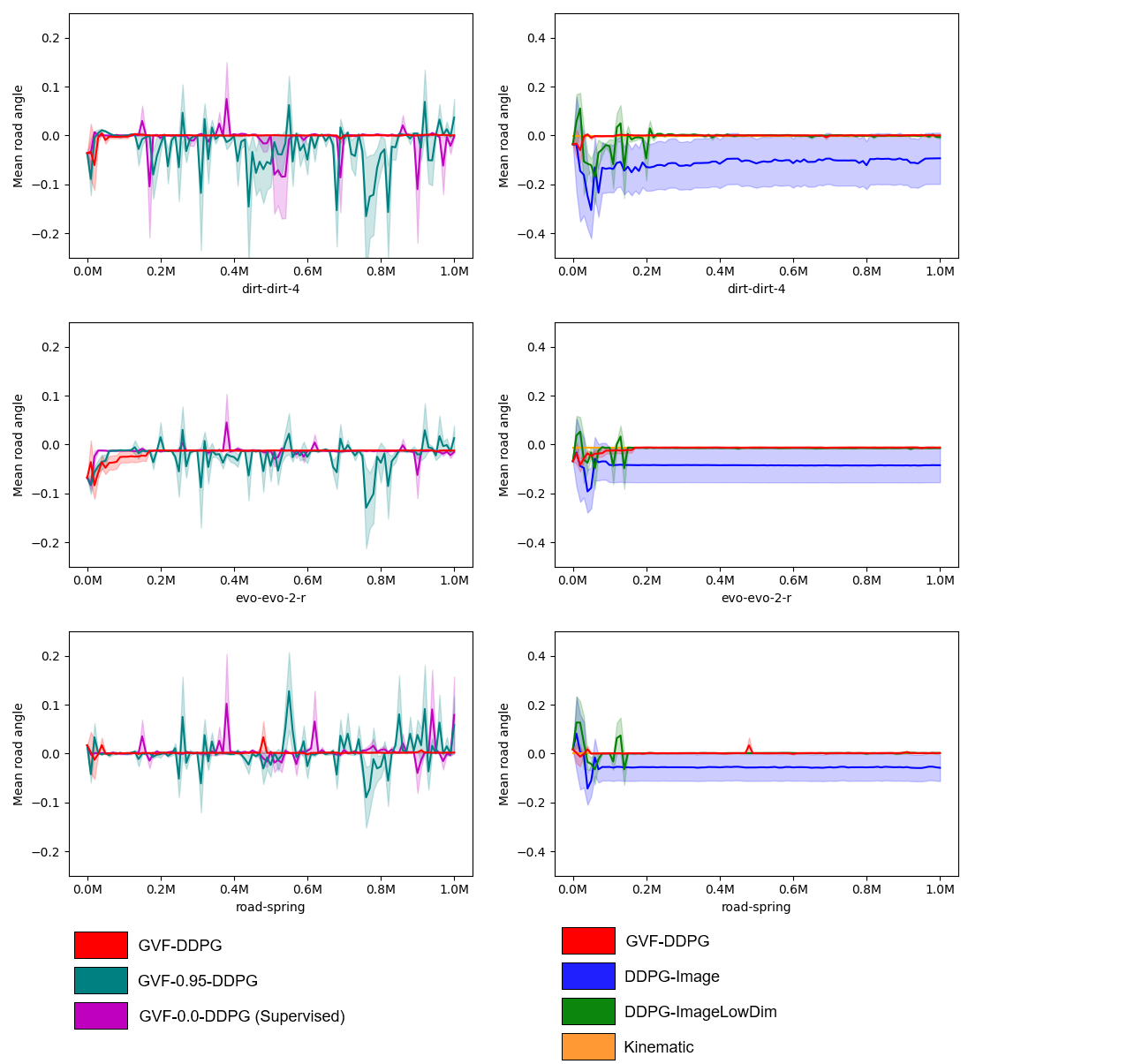}
	\caption{Mean road angle during training for dirt-dirt-4, evo-evo-2 and road-spring}
	\label{fig_road_angle}
\end{figure}

\begin{figure}[!h]
	\centering
	\includegraphics[width=12cm]{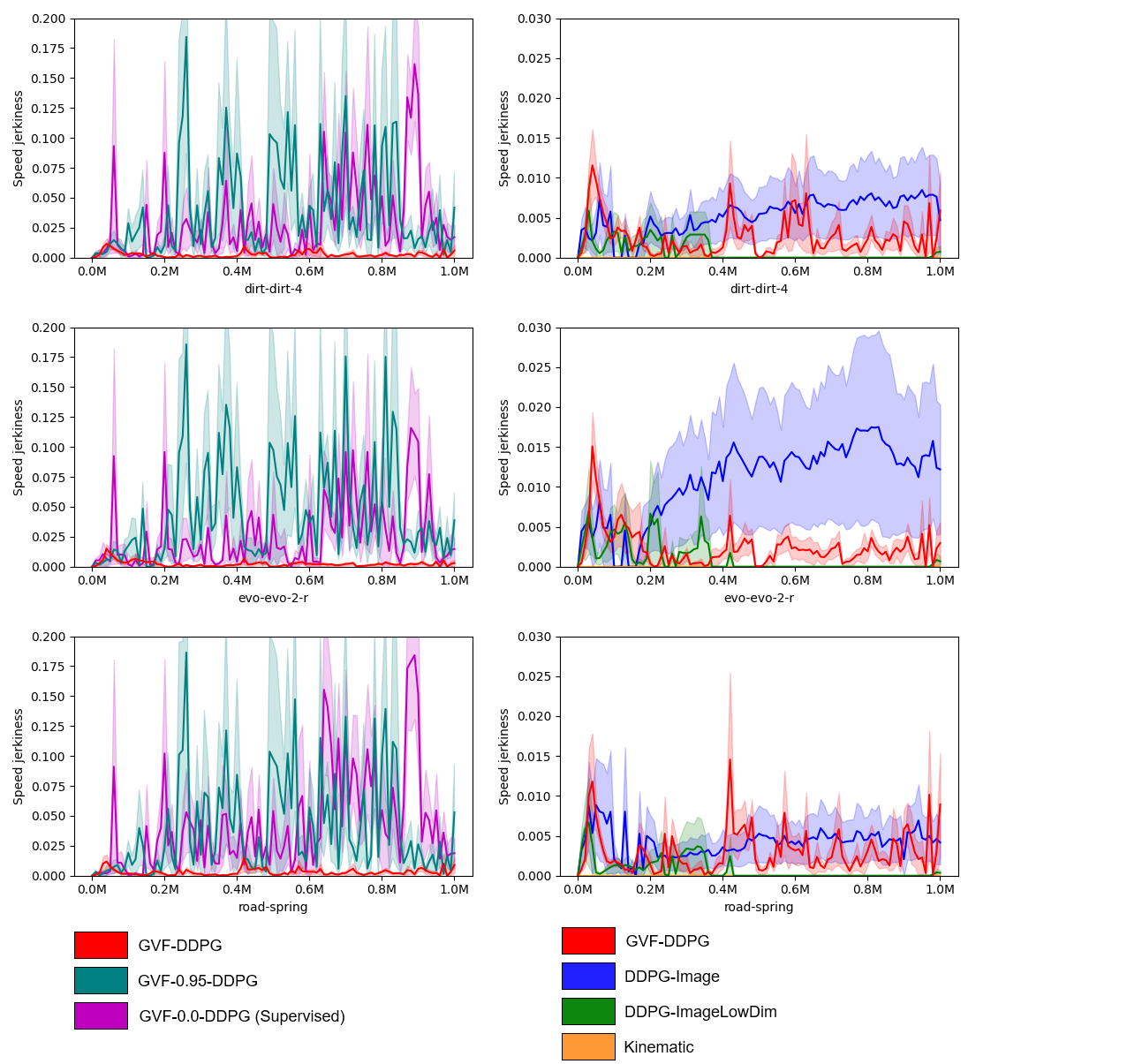}
	\caption{Standard deviation of the change in target speed action during training for dirt-dirt-4, evo-evo-2 and road-spring}
	\label{fig_delta_action_speed_std}
\end{figure}

\begin{figure}[!h]
	\centering
	\includegraphics[width=12cm]{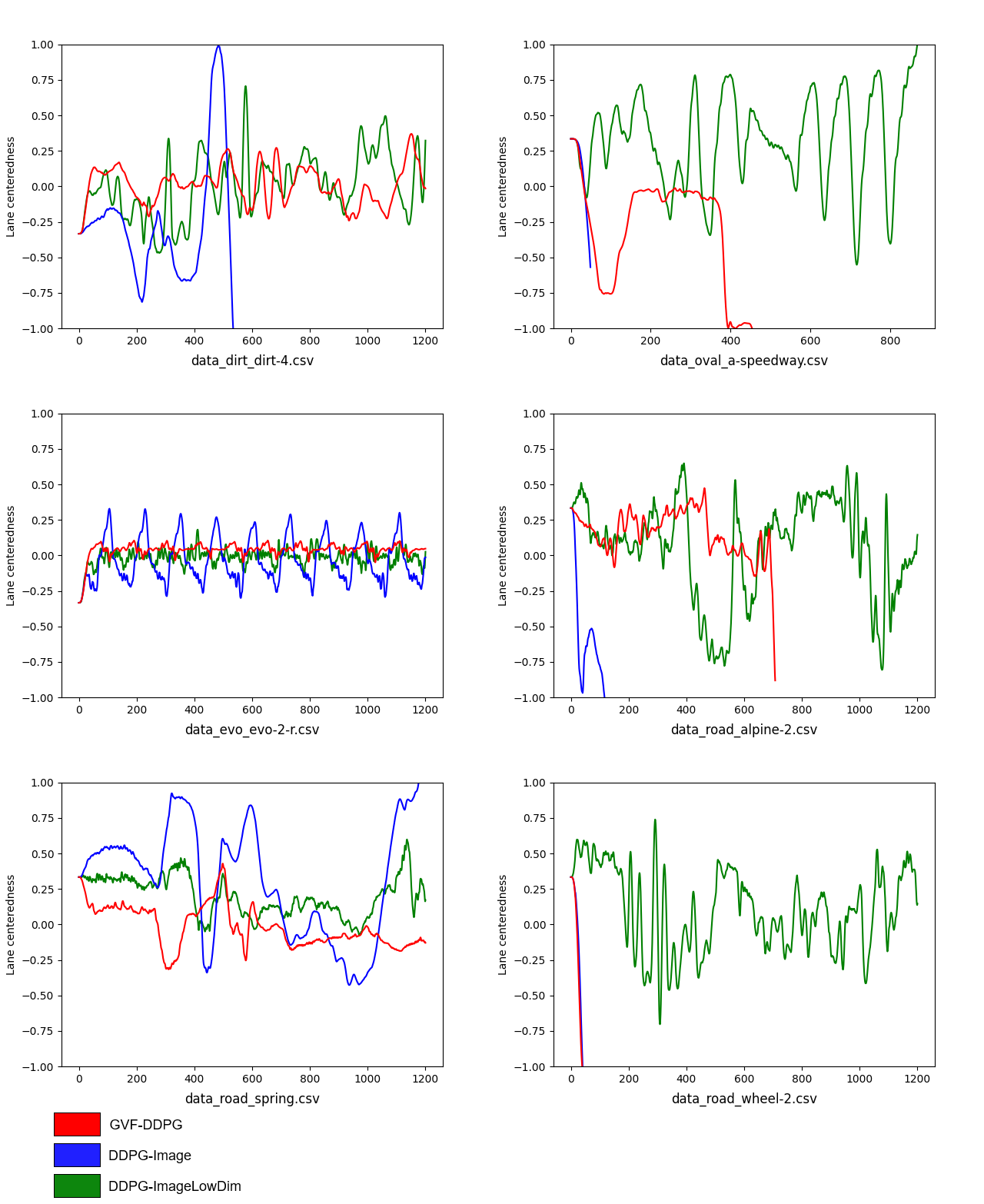}
	\caption{The lane centeredness position on the (a) alpine-2, (b) evo-2-r, (c) dirt-4, (d) wheel-2, (e) spring, and (f) a-speedway roads in TORCS.}
\label{fig_track_lane_centeredness}
\end{figure}

The experimental results were averaged over 5 runs and mean and standard deviations plotted based on performance measured on the test roads during training.
The learning curves for the action-value networks for each of the DDPG agents, as well as learning curves for GVF predictions and behavior estimation for GVF-DDPG are shown in Figure \ref{fig_learning_curve}.
The average episode length is shown in Figure \ref{fig_episode_length}.
The average lane centeredness and road angle during each episode are plotted in Figures \ref{fig_road_centeredness} and \ref{fig_road_angle}.
We can see how the GVF-DDPG with predictions over multiple time scales is able to maintain lane centeredness and road angle better than GVF-DDPG with myopic prediction ($\gamma=0.0$) and future predictions with only $\gamma=0.95$.
The average lane centeredness and road angle is not substancially different though amoung the learned methods; however, DDPG-Image struggles a bit largely due to instability in learning as the high variance is due to some failed runs.
Figure \ref{fig_delta_action_speed_std} shows the standard deviation in the change in the target speed action at each time step across an episode on each test road; this measures the jerkiness of the speed controller.
GVF-DDPG and DDPG-LowDim are both able to control speed comfortably.
Finally, Figure \ref{fig_track_lane_centeredness} shows the lane centeredness on all six of the test roads during a final evaluation after training was completed.
All six roads in the test set were challenging but the test roads a-speedway, alpine-2, and wheel-2 were especially challenging because the image of the roads were too different from the training roads.
Nevertheless, on the dirt-4, evo-2-r, and spring roads, the lane centeredness of the the methods was quite good for all learned methods except for DDPG-Image.
Note that while DDPG-LowDim performs well in learning to steer and control speed with lane centeredness and road angle, it required that information to learn to steer the vehicle which may be expensive or prohibitive to obtain in all situations such as in GPS-denied locations or locations where there is no map or it is out of date.

\clearpage
\section{Jackal Robot Experiments}

The Jackal robot is equipped with a 5MP camera using a wide angle lens, an IMU sensor and an indoor Hokuyo UTM-30LX LIDAR sensor with a $270^{\circ}$ scanning range and 0.1 to 10 meters scanning distance.
The objective is to drive the robot using a camera in the center of a lane marked with tape using only data collected in the real-world.
This is a challenging problem for reinforcement learning algorithms as RL, especially model-free, usually takes a prohibitive amount of time and samples to learn.
Training directly from real world experience requires addressing multiple issues such as minimizing wear and tear on the robot over long period of interaction, and the need of human supervision during training in order to prevent robot crashes and recharge the battery.

There are two learned controllers, called GVF and E2E respectively, and one classical baseline called MPC (model predictive control).
The learned controllers outputs a steering angle $a_t^{steer}$ and target speed $a_t^{speed}$ based on the image taken by the camera in order drive centered in a closed loop track on a carpeted floor.
The MPC outputs a steering angle $a_t^{steer}$ and target speed $a_t^{speed}$ based on localization of the robot in a map constructed of the environment to follow a sequence of waypoints.
Localization, map and waypoints are needed to train the GVF controller; however, this information is no longer used during testing.

\subsection{Training and testing environment}

The environment used for collecting data and evaluating the agents included two types of carpet floor - each with a different amount of friction.
The evaluation roads were done on one carpet only which was included in only about 20\% of the training data; the rest of the training data was on the other type of carpet to provide a generalization challenge for our learned controllers.
The friction was quite high on both carpets and it caused the agent to shake the camera violently while turning since the robot employs differential steering; tape on the wheels helped reduce the friction a bit.
Nevertheless, localization techniques using wheel odometry was deemed unsuitable and LIDAR-based localization was used instead.
LIDAR localization was not highly accurate but was sufficient; preliminary tests showed that it
was repeatable to within roughly 5 centimeters.

Nine closed loop roads were created by placing left and right lanes markings on the carpeted floor separated to form a consistent lane width of roughly 76 centimeters for all the roads.
6 of the roads were selected for collecting data to train our learned agents.
Data was collected in both directions.
The remaining 3 roads were reserved for testing the performance of the policies.
The poses and orientations of a sequence of waypoints were established to denote the center of lane which is the desired path of the agent; this was used to train our agents on the training roads and evaluate them on the test roads.
The center waypoints were collected by an expert manually and carefully navigating the Jackal robot on the road in the approximate center of lane; while this was inaccurate, it was not found to harm learning since the GVF approach was able to generalize and learn features to drive the vehicle centered in the lane.
The LIDAR-based localization produced poses periodically to form the center waypoints; these were cleaned up by removing overlapping waypoints to form a closed loop path.
However, it did not produce poses at a consistent sampling frequency and thus a linear interpolation method was used to fill in the gaps and provide localization and center waypoint information at every time step.
The purpose of the center waypoints was to compute the road angle and lane centeredness of the robot in the road at any given time which is needed to train the GVF predictions and evaluate all of our controllers.

\begin{figure}[!h]
	\centering
	\includegraphics[width=12cm]{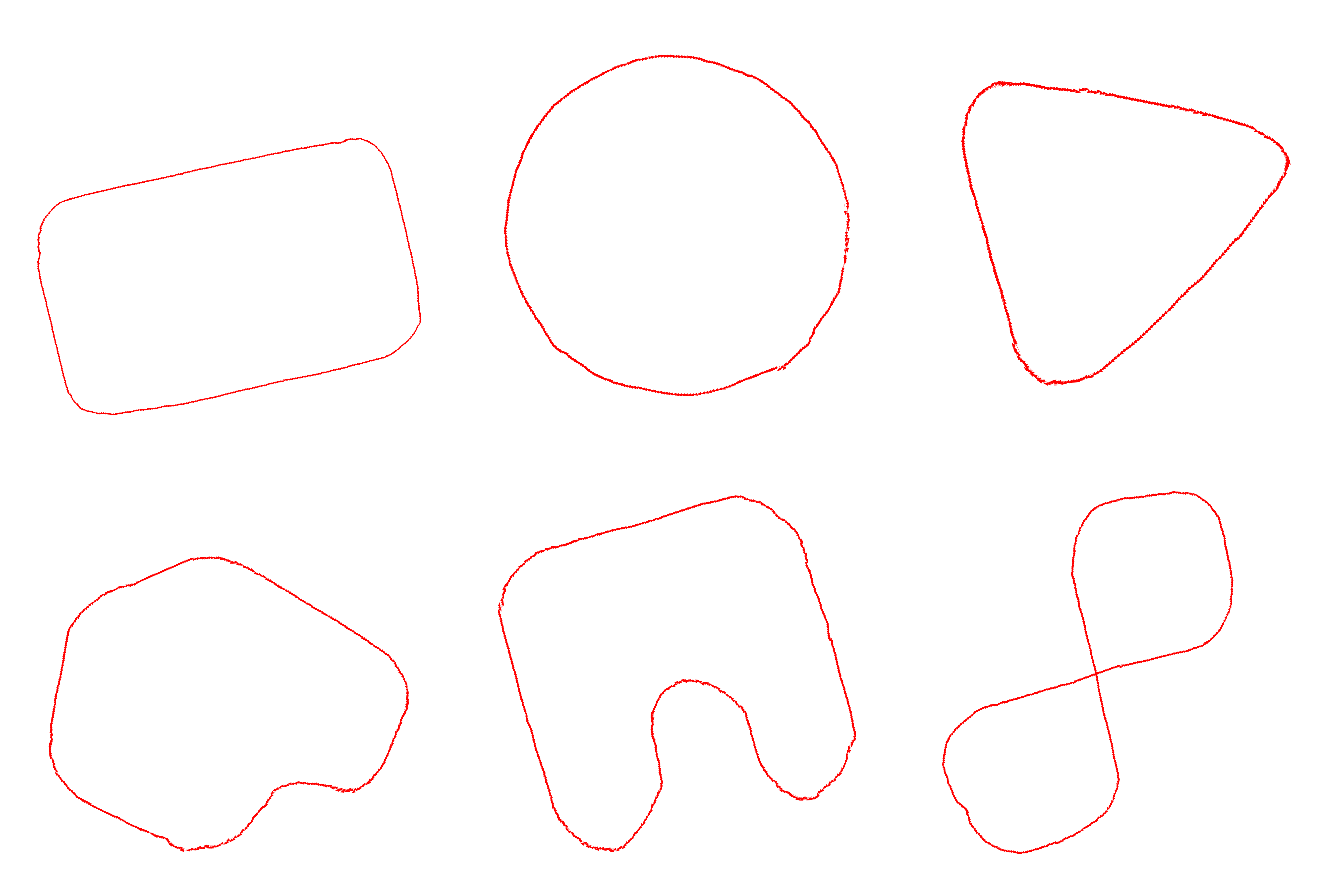}
	\caption{Six roads for training.  The second row shows more complex road structure.  The rectangle road has rectangular edges at all corners.}
	\label{fig_jackal_train_tracks}
\end{figure}

\begin{figure}[!h]
	\centering
	\includegraphics[width=12cm]{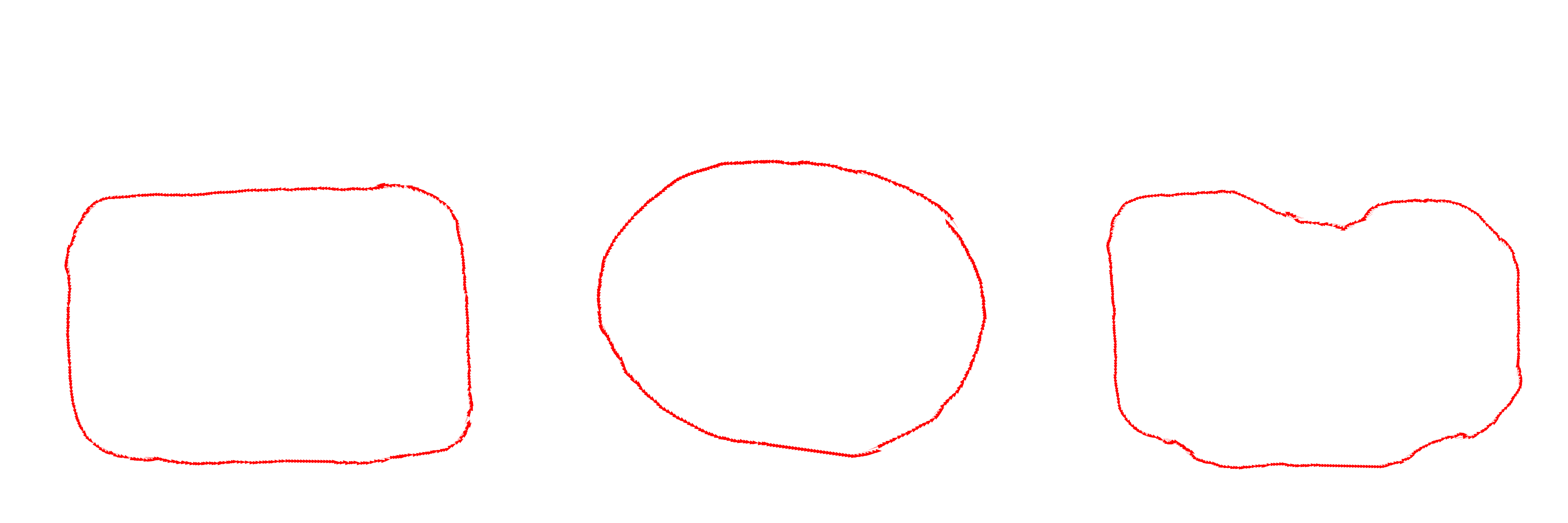}
	\caption{Three roads for testing.  Left to right: (1) rectangle with rounded corners; (2) oval; (3) complex.}
	\label{fig_jackal_test_tracks}
\end{figure}

\begin{figure}[!h]
	\centering
	\includegraphics[width=12cm]{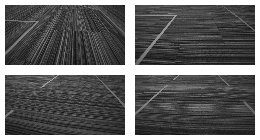}
	\caption{Input images of normal lane markings (top row) and damaged lane markers (bottom row).}
	\label{fig_jackal_train_tracks_camera}
\end{figure}

The center waypoints for the training and testing roads are depicted in Figure \ref{fig_jackal_train_tracks} and Figure \ref{fig_jackal_test_tracks}, respectively.
The first three training roads were simple geometric shapes while the other three were a bit more complex.
The first test road was the most similar to the training data where the outer edge of the four corners were rounded.
The second test road was an oval shape to evaluate how well the agent maintained a continuous turn.
The third test road was a complex shape with multiple sudden turns not in the training data.
This tests generalization to new roads.
All methods were evaluated at 0.25 m/s and 0.4 m/s maximum speeds and clock-wise (CW) and counter clock-wise (CCW) directions.
In order to test robustness, some test roads were altered by degrading or damaging.
An example of an image received from the robot with and without damage to the lane markers is shown in Figure \ref{fig_jackal_train_tracks_camera}.

\subsection{Data collection}
Both learned agents -- GVF and E2E -- were trained with batch reinforcement learning \citep{fujimoto2018off} where the GVF method learned a predictive representation.
Rosbags were collected with the Jackal robot each containing around 30 minutes to 2 hours of data for a total of 40 hours of data containing approximately 1.5 million images.
The camera image, localization pose, IMU measurement, and action taken by the controller was recorded at each time step.
The Jackal robot was controlled using a random walk controller that was confined to the road area to provide sufficient and safe exploration.
The map was created with Hector SLAM \citep{KohlbrecherMeyerStrykKlingaufFlexibleSlamSystem2011} and the localization produced by Adaptive Monte-Carlo Localization \cite{thrun2002probabilistic}.
This localization method was prone to a significant amount of noise.

The random walk controller was based on a pure pursuit controller, where action taken at each time step is defined by
\begin{equation}
\begin{split}
a_t^{steer} & = \text{clip}(\text{angle}(p_t, p^*_{k(t)}) - \theta^z_t, -\pi / 2, \pi / 2) \\
a_t^{speed} & = \text{clip}(v^*_{k(t)}, 0.2, 0.5)
\end{split}
\label{eq_jackal_pure_pursuit_actions}
\end{equation}
where $\theta^z_t$ and $p_t$ were the yaw angle and the 2-dimensional position of the robot in the real world (obtained from localization), $p^*_{k(t)}$ and $v^*_{k(t)}$ were the target position and linear velocity at time $t$, clip($x$, MinVal, MaxVal) is the clip function that operates in scalar and vector element-wise, and angle$(p_t, p^*_{k(t)})$ is the function that returns the yaw angle in the real world of a vector that points from $p_t$ to $p^*_{k(t)}$.
The target position $p^*_{k(t)}$ and linear velocity $v^*_{k(t)}$ were encapsulated in the next target pose at index $k(t)$ in the sequence of target poses:
\begin{equation}
\begin{split}
k(1) & = 1 \\
k(t+1) & =    
	\begin{cases}
      k(t) + 1 \text{ if } ||p_t-p^*_{k(t)}||_2 < 0.025 \\
      k(t) \text{ otherwise }
	\end{cases}
\end{split}
\label{eq_jackal_waypoint_update}
\end{equation}
Thus, the robot advanced to the next target position and linear velocity in the target pose sequence once it arrived within 2.5 centimeters of the current target position.
In order to provide efficient and safe exploration that can be confined to the road area, the target position $p^*_j$ was based on the position $\tilde{p}_j$ of the center waypoints collected earlier with some noise added:
\begin{equation}
\begin{split}
p^*_{j} & = \tilde{p}_{j \% N} + \varepsilon^p_j \\
v^*_{j} & = 
	\begin{cases}
		v^*_{j-1} + \varepsilon^v_j \text{ if } j > 1 \\
		0.35 \text{ if } j = 1
	\end{cases}
\end{split}
\label{eq_jackal_target_noise}
\end{equation}
where $N$ is the number of points that define the center waypoints of the closed loop road.
$\varepsilon^p_j$ and $\varepsilon^v_j$ were the noises added at each time step:
\begin{equation}
\begin{split}
	\varepsilon^p_j & = 
		\begin{cases}
			\text{clip}(\varepsilon^p_{j-1} + \mathcal{N}(0, 0.02 * \mathbbm{1}), -0.3, 0.3) \text{ if } j > 1 \\
			[0, 0]^\intercal \text{ if } j = 1
		\end{cases} \\
	\varepsilon^v_j & = \mathcal{N}(0, 0.02)
\end{split}
\label{eq_jackal_exploration_noise}
\end{equation}
The noises for the poses were clipped so that the robot would not explore too far outside the road area.

The rosbags were processed to synchronize the sensor data streams at a fixed sample frequency of 10Hz and compute the lane centeredness $\alpha_t$, road angle $\beta_t$, and speed $v_t$ of the robot at each time step:
\begin{equation}
\begin{split}
	\nu_t & = \text{knn}(p_t, S_t) \\
	\alpha_t & = \text{clip}(\frac{||p_t - \nu_t||_2}{H}, -1.0, 1.0) \\
	\beta_t & = \text{clip}(\text{angle}(p_t, \nu_t) - \theta_t^z, -\pi/2, \pi/2) \\
\end{split}
\end{equation}
where $\text{knn}(x, S)$ returns $\nu$ as the closest point to $x$ in $S$ using k-nearest neighbor and $H=38$ centimeters as the half lane width.
$S_t$ is a pruned set of center waypoints  where $S_t = \{\tilde{p}_t\}$ for all roads, except for the figure 8 road in the lower right of Figure \ref{fig_jackal_train_tracks} where $S_t$ was based on a sliding window:
\begin{equation}
\begin{split}
	S_t = 
		\begin{cases}
			\{\tilde{p}_t\} \text{ if } j=1 \\
			\{\tilde{p}_{t=I_{t-1}-w \rightarrow I_{t-1}+w}\} \text{ if } j > 1
		\end{cases}
\end{split}
\end{equation}
where $I_{j-1}$ is the index of $\nu_{t-1}$ in $S_{t-1}$ at the previous time step and $w=10$ is the size of the sliding window.
Negative indices are wrapped to the beginning of the road.
The speed was estimated using the change in position over a single time step which was quite noisy but more reliable than speed returned from the robot's odometry.
Due to computation constraints on the robot, the localization messages were output at less than 10Hz; thus, a linear interpolation was used to fill in missing poses and orientations in the data and synchronize the data streams.

Images from camera with original size $1920 \times 1080$ were cropped to $960 \times 540$ region in the center and then additionally top-cropped by 60.
The images were then downsampled by a factor of 8 in both spatial dimensions to give a final size of $120 \times 60$ and then converted to gray scale.
To improve generalization in deep learning and balance the left-right biases in the data, augmented data was created with horizontally flipped images along with the corresponding signs of the lane centeredness $\alpha_t$, road angle $\beta_t$, and steering action $a_t^{steer}$) flipped.

\subsection{GVF-BCQ Training}
The predictive neural network used was identical to the one used in the GVF-DDPG TORCS experiments.
The offline version of predictive learning, cf. algorithm \ref{alg_gvf_offpolicy_train_without_mu_offline}, was used to train the GVFs.
The transitions in the data were loaded into a replay buffer in the same order that the transitions were observed in the rosbag where mini-batches of size 128 where sampled from the growing replay buffer and used to update the GVFs.
The GVFs were updated for 5 million steps followed by BCQ for an additional 5 million steps.
The order that the rosbags were loaded into the replay buffer was randomized.
The replay buffer had a maximum capacity of 0.5 million samples; once the replay buffer was filled, the oldest samples were removed.
Training began once the replay buffer reached 0.1 million samples.
While a better approach would have been to sample mini-batches from the entire data set from the beginning, our approach was found to be effective and required minimal changes to the data loader.

\begin{figure}[!h]
	\centering
	\includegraphics[width=12cm]{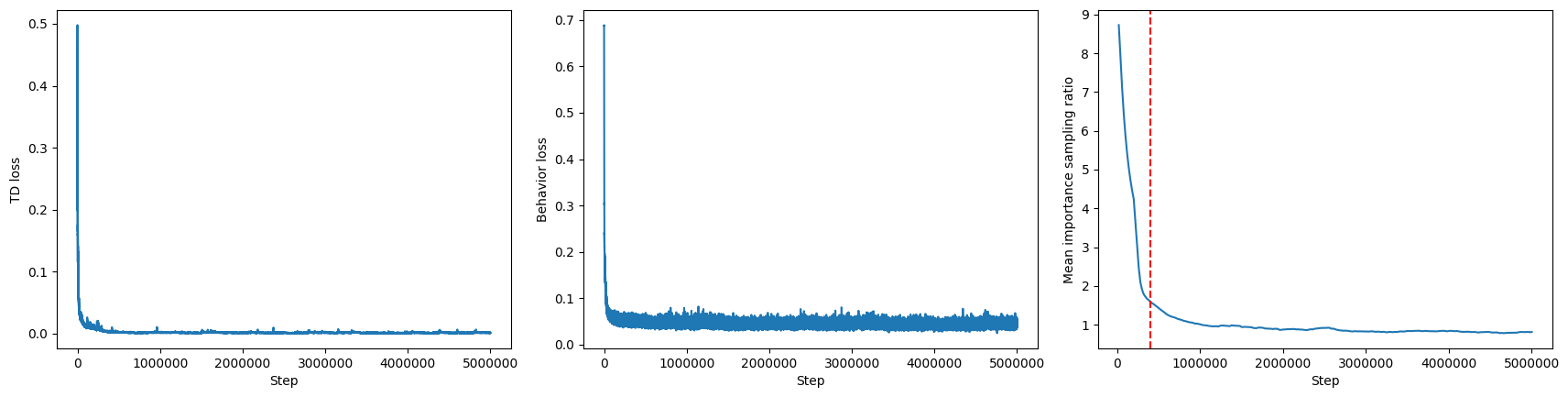}
	\caption{TD loss, behavior loss and mean importance sampling ratio in the buffer over training steps. Red vertical dash line is the point when the buffer is full.}
	\label{fig_jackal+is_ratio}
\end{figure}

Estimating the behavior distribution for the GVF predictions was done in the same manner as with TORCS.
$\eta(a|s)$ was a uniform distribution defined on the interval $[-\pi/2,\pi/2]$ for the steering action and a uniform distribution defined on the interval $[0, 1]$ for the target speed action.
The BCQ VAE model was a fully connected network receiving $(d_a + d_s)$ inputs that fed into a hidden layer of 256 and into an encoder output layer $d_z$ containing the latent variables.
The decoder of the VAE received the encoded latent variables $d_z$ that fed into a hidden layer of 256 and output the reconstructed action of size $d_a$.
The actor network was fully connected and received a vector of size $(d_a + d_s)$, where $s$ was the predictive representation of size $d_s$ and the action $a_t$ generated by the VAE, that fed into a hidden layer of size 256 and output a perturbation to the action of dimension $d_a$.
The critic  received a vector of size $(d_a + d_s)$ that fed into a hidden layer of size 256 and output a scale representing the value of the state and action.
The action dimension was $d_a=2$, the predictive representation was of dimension $d_s=11$(consisting 8 predictions + last steering action + last target speed action + last robot speed) and the latent vector dimension was $d_z=4$ which was a Normal distribution parameterised by mean and log standard deviation. 
All networks used ReLU activation for the hidden layers and linear activation for the outputs.
Just as with BCQ, the actor outputs an action of dimension $d_a$ dimension that is a perturbation to the action reconstructed through the VAE; the two actions are combined to produce the final action of size $d_a$ such that the batch-constrained property is satisfied.
The action output from the actor and VAE were clipped to $[-\pi/2, \pi/2]$ for steering and $[0.1, 0.6]$ for the target speed.
The weight of the KL divergence loss was 0.5.
The learning rate was $10^{-4}$ for both GVF and BCQ model training.
Figure \ref{fig_jackal+is_ratio} shows the training curves for the predictive representation (GVFs) including the temporal-difference loss, behavior model loss and mean importance sampling ratio.

\subsection{End-to-end BCQ Baseline Training}
Nearly the same training setup used for the GVF method was applied to the E2E method.
The hyperparameters, training settings, activation functions for the output and action clipping are exactly the same as GVF, except for the network architectures.
The encoding branch of the VAE, the actor and critic networks use the same architecture as the prediction model of GVF-DDPG.
Last steering, last target speed action and last robot speed are merged into flattened output from the convolutional layers the same way as in DDPG-LowDim.

For a fair comparison, E2E with BCQ was trained for 10 millions update steps - the same number as GVF and BCQ combined in our GVF method.
The agent was then tested on the rectangle test road and it was found that E2E performed very poorly. 
The agent was too slow reaching an average speed of about 0.18 m/s whereas the GVF method was able to reach double that speed.
In addition, E2E steered very poorly and was often not centered in the lane; unlike the GVF method which was observed to be quite robust, the E2E method sometimes drove out of the lane where an emergency stop was needed to prevent collision.
For this reason, E2E was only compared to GVF on the rectangle test road; and we focused on comparisons against the MPC controller which was more robust than E2E.
A detailed evaluation of E2E is shown in Figure \ref{fig_jackal_image_bcq} and Table \ref{table_summary_jackal_result_gvf_vs_e2e}.

\begin{figure}[!h]
	\centering
	\includegraphics[width=\textwidth]{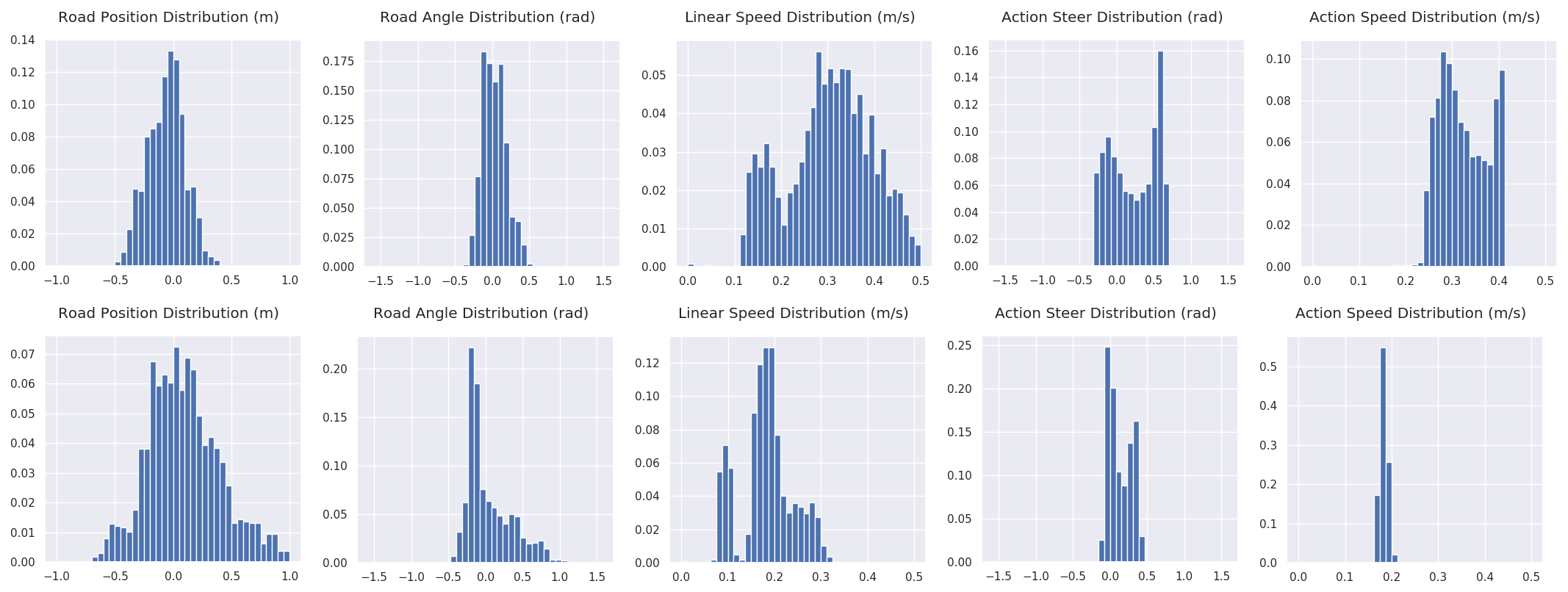}
	\caption{Distribution of lane centeredness, road angle, speed and action distribution of GVF and E2E on the rectangle test track at 0.4 speed, counterclockwise direction.}
	\label{fig_jackal_gvf_vs_e2e}
\end{figure}

\subsection{MPC Baseline}
An MPC baseline using standard ros nodes for the Jackal robot were used for controlling the Jackal robot.
The baseline was tuned for 0.4 m/s; however, it was challenging to achieve good performance due to limited computation power for the look ahead and inaccurate modeling of the steering characteristics on carpet floors.
The best performance was achieved for 0.25 m/s but significant oscillation was observed for 0.4 m/s that was challenging to completely eliminate.
The center waypoints provided as input to the MPC controller were processed so that the minimum distance between two consecutive waypoints was 2.5cm; the waypoints were then downsampled by a factor of 16 in order to increase their separation and increase the look ahead distance; different downsampling factors was tested but oscillation was never completely eliminated.
The MPC had an optimization window of 5 steps into the future.
This ensured the MPC look ahead was far enough into the future while also maintaining a reasonable computation burden needed for real time control.

\subsection{Test Results}

A comparison of GVF and E2E is given in Tables \ref{table_jackal_result_gvf_vs_e2e1} and \ref{table_jackal_result_gvf_vs_e2e2}.
This section provides a detailed comparison of GVF and MPC methods at different speeds and directions.
For the GVF method, the agent was trained to drive as fast as possible (i.e. no target speed was given) and the speed command was clipped at the maximum target speed.
In our evaluation at 0.25 m/s in the counterclockwise direction, the two controllers performed similarly; the primary differences in the methods were observed at higher speeds demonstrating that the GVF method was able to scale better than the MPC.
The controllers started at the same position and heading angle and they were allowed to run for exactly 300 seconds.
The agents were evaluated based on the following criteria:

\begin{itemize}
\item Reward per second: $\frac{1}{N} \sum_{t=1}^N r_t$
\item Average speed: $\frac{1}{N} \sum_{t=1}^N v_t$
\item Average absolute lane centeredness: $\frac{1}{N} \sum_{t=1}^N |\alpha_t|$
\item Average absolute road angle: $\frac{1}{N} \sum_{t=1}^N |\beta_t|$
\item Near out of lane\footnote{ratio of time steps where the agent's absolute lane centeredness is greater than 0.75}:  $\frac{1}{N} \sum_{t=1}^N \mathbbm{1}_{|\alpha_t| > 0.75}$\footnote{Where $\mathbbm{1}$ is the indicator function.}.
\item First Order Comfort\footnote{First order comfort is the negative absolute change of action taken by the agent in one time step.  Higher comfort scores are better.  Both steering and speed actions considered separately.}: $-\frac{1}{N-1} \sum_{t=1}^{N-1} |a_{t+1}-a_t|$
\item Second Order Comfort\footnote{Second order comfort is the negative absolute change of the first order comfort in one time step.  Higher comfort scores are better.  Both steering and speed actions considered separately.}: $-\frac{1}{N-2} \sum_{t=1}^{N-2} |(a_{t+2}-a_{t+1})-(a_{t+1}-a_t)|$
\end{itemize}

\begin{figure}[!h]
	\centering
	\includegraphics[width=\textwidth]{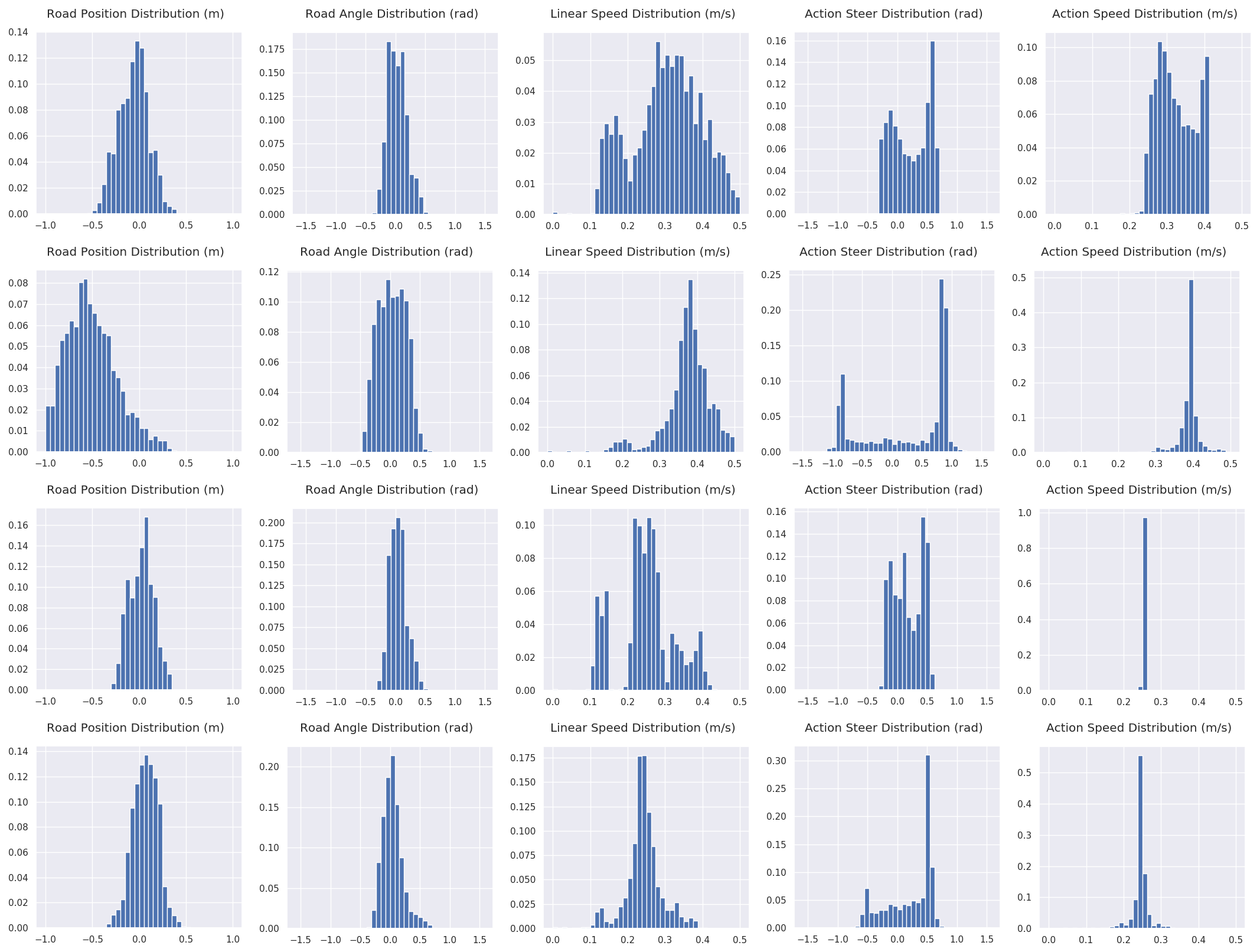}
	\caption{Distribution of lane centeredness, road angle, speed and action distribution of GVF-BCQ and MPC at 0.4 m/s and 0.25 m/s on the rectangle test track. From top to bottom: GVF-0.4-CCW, MPC-0.4-CCW, GVF-0.25-CCW, MPC-0.4-CCW\protect\footnotemark}
	\label{fig_jackal_gvf_vs_mpc}
\end{figure}

\begin{figure}[!h]
	\centering
	\includegraphics[width=\textwidth]{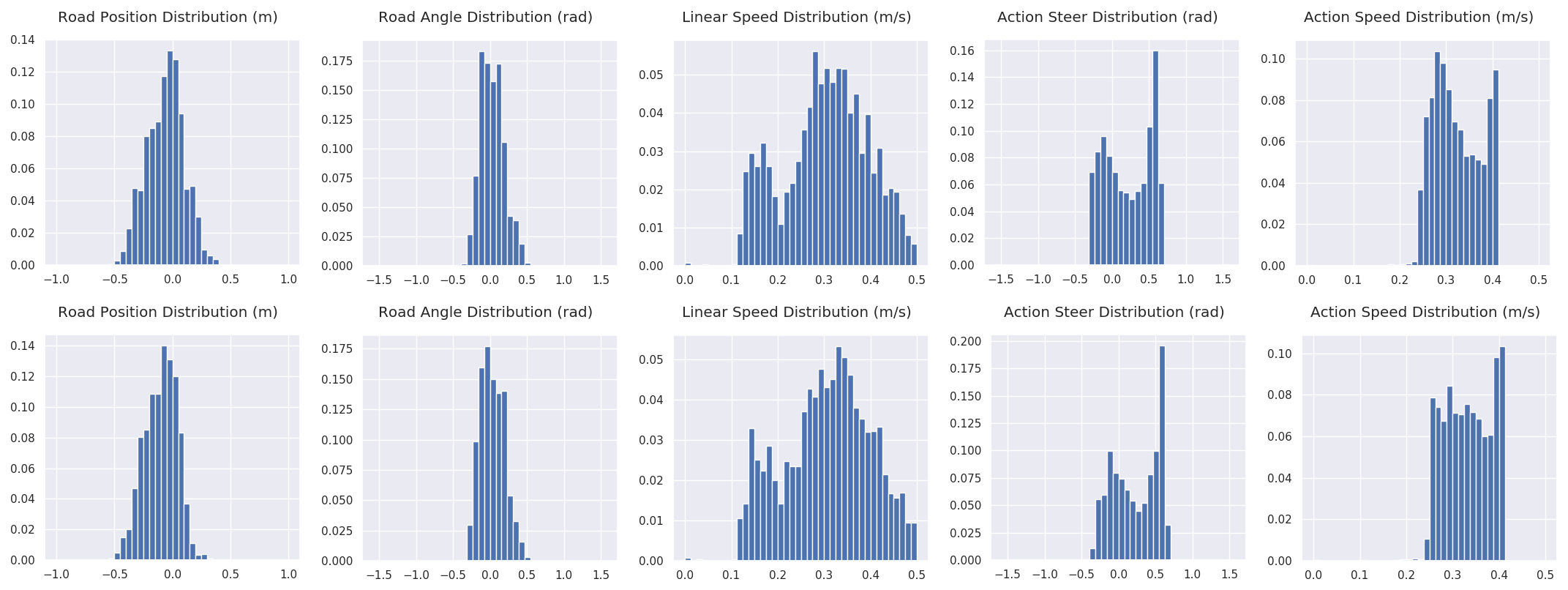}
	\caption{Distribution of lane centeredness, road angle, speed and action distribution on the rectangle test road. From top to bottom: GVF-BCQ-0.4, GVF-BCQ-0.4 with lane marking damage on the rectangle road. The similarities highlight the robustness of GVF-BCQ to the introduction of damaged lanes.}
	\label{fig_jackal_gvf_N_vs_D}
\end{figure}

\begin{figure}[!h]
	\centering
	\includegraphics[width=\textwidth]{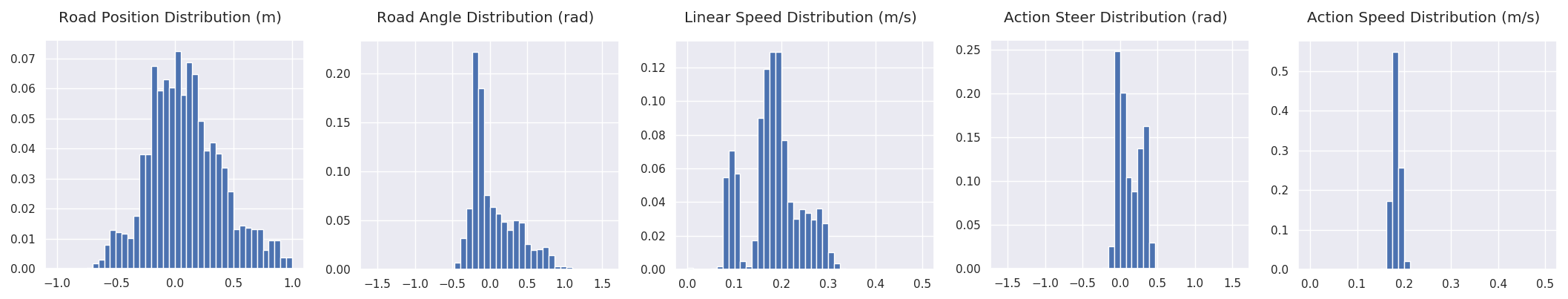}
	\caption{Distribution of lane centeredness, road angle, speed and action distribution of E2E-BCQ on the rectangle test road at 0.4 speed, counterclockwise direction}
	\label{fig_jackal_image_bcq}
\end{figure}

In order to provide more insight into the performance of the controllers, we also investigated the distributions of $\alpha_t$, $\beta_t$, $v_t$ and $a_t$.
Details evaluations are shown in Tables \ref{table_jackal_result_gvf_vs_mpc1} and \ref{table_jackal_result_gvf_vs_mpc2}.
Experiments are named according to the method used, the selected target speed and the direction of the road loop (i.e. counter-clock-wise versus clock-wise).
For example, GVF-0.4-CCW points to the test of the GVF controller with 0.4 m/s target speed in the counter-clock-wise direction.
Additionally, experiments on roads with damaged lane markings are denoted with suffix -D.
It can be seen from Tables \ref{table_jackal_result_gvf_vs_mpc1} and \ref{table_jackal_result_gvf_vs_mpc2} that the GVF method beat the MPC in reward and was better on all tracks at both speed values without access to localization information during testing.
The reason for this can be explained by looking at the average lane centeredness of the agent.
The MPC performed as well as the GVF method while maintaining good speed; however, it fails at keeping the vehicle in the center of the lane.
The reason may be due to a number of different factors including possible inaccuracies in the MPC forward model resulting from the friction between the wheels and the carpet in the test runs, especially at higher speeds.
The MPC suffered from many near out of lane events and had trouble staying within the lane markings of the oval road.
The GVF method was better at controlling steering, leading to much higher average reward even though it had lower average speed at 0.4 m/s max speed.
Additionally, the GVF method was much better in achieving smooth control.
These points are reflected in Figure \ref{fig_jackal_gvf_vs_mpc} on the rectangle test track where the MPC lane centeredness distribution is skewed to one side and its steering action distribution has two modes that are far from zero while the GVF method distributions are more concentrated around zero.
Finally, the GVF method generalized well to damaged lane markings and distractions in the visual images as shown in the similar scores and similar distributions in Figure \ref{fig_jackal_gvf_N_vs_D} and Table \ref{table_jackal_result_N_vs_D1} and \ref{table_jackal_result_N_vs_D2}.

\footnotetext{Note that measured vehicle speed might not be equal to speed action from the agent due to physical constraints of the environment and noises in measurement.}

\begin{table}
\centering
\caption{Comparison of GVF method and MPC on all the test roads at different speeds and directions}
\label{table_jackal_result_gvf_vs_mpc1}
\begin{tabular}{|l|l|l|l|l|l|l|} 
\hline
                                                                              & Experiment & \multicolumn{1}{c|}{\begin{tabular}[c]{@{}c@{}}Reward\\ per \\ second \textuparrow \end{tabular}} & \multicolumn{1}{c|}{\begin{tabular}[c]{@{}c@{}}Average\\ speed \textuparrow \end{tabular}} & \multicolumn{1}{c|}{\begin{tabular}[c]{@{}c@{}}Average \\ offcentered\\ -ness \textdownarrow\\ (normalized) \end{tabular}} & \multicolumn{1}{c|}{\begin{tabular}[c]{@{}c@{}}Average \\ absolute\\ road angle \textdownarrow \end{tabular}} & \multicolumn{1}{c|}{\begin{tabular}[c]{@{}c@{}}Near \\ out\\ of lane \textdownarrow \end{tabular}}  \\ 
\hline
\multirow{6}{*}{\begin{tabular}[c]{@{}l@{}}Rectangle\\ shape \end{tabular}}   & GVF-0.4-CCW  & \textbf{2.6835}                                                                        & 0.3205                                                                          & \textbf{0.1345}                                                                                               & \textbf{0.1315}                                                                                  & \textbf{0.0}                                                                           \\
                                                                              & MPC-0.4-CCW  & 0.9700                                                                                 & \textbf{0.3833}                                                                 & 0.5252                                                                                                        & 0.1943                                                                                           & 0.2042                                                                                 \\ 
\cline{2-7}
                                                                              & GVF-0.4-CW   & \textbf{2.2915}                                                                        & 0.3140                                                                          & \textbf{0.2217}                                                                                               & \textbf{0.1586}                                                                                  & \textbf{0.0}                                                                           \\
                                                                              & MPC-0.4-CW   & 0.1282                                                                                 & \textbf{0.3836}                                                                 & 0.9086                                                                                                        & 0.1916                                                                                           & 0.6786                                                                                 \\ 
\cline{2-7}
                                                                              & GVF-0.25-CCW & \textbf{2.1442}                                                                        & \textbf{0.2467}                                                                 & \textbf{0.1098}                                                                                               & \textbf{0.1181}                                                                                  & 0.0                                                                                    \\
                                                                              & MPC-0.25-CCW & 1.1971                                                                                 & 0.2412                                                                          & 0.1218                                                                                                        & 0.1308                                                                                           & 0.0                                                                                    \\ 
\hline
\multirow{6}{*}{\begin{tabular}[c]{@{}l@{}}Oval\\ shape \end{tabular}}        & GVF-0.4-CCW  & \textbf{2.4046}                                                                        & 0.3501                                                                          & \textbf{0.2754}                                                                                               & 0.2125                                                                                           & \textbf{0.0145}                                                                        \\
                                                                              & MPC-0.4-CCW  & 0.8928                                                                                 & \textbf{0.3825}                                                                 & 0.5293                                                                                                        & \textbf{0.1963}                                                                                  & 0.2275                                                                                 \\ 
\cline{2-7}
                                                                              & GVF-0.4-CW   & \textbf{2.4848}                                                                        & 0.3658                                                                          & \textbf{0.2953}                                                                                               & \textbf{0.1922}                                                                                  & \textbf{0.0}                                                                           \\
                                                                              & MPC-0.4-CW   & -0.7168                                                                                & \textbf{0.3836}                                                                 & 1.3182                                                                                                        & 0.2095                                                                                           & 0.9122                                                                                 \\ 
\cline{2-7}
                                                                              & GVF-0.25-CCW & \textbf{1.5112}                                                                        & \textbf{0.2473}                                                                 & \textbf{0.3645}                                                                                               & 0.1466                                                                                           & \textbf{0.0332}                                                                        \\
                                                                              & MPC-0.25-CCW & 0.0225                                                                                 & 0.2296                                                                          & 0.9565                                                                                                        & \textbf{0.1381}                                                                                  & 0.8792                                                                                 \\ 
\hline
\multirow{6}{*}{\begin{tabular}[c]{@{}l@{}}Complex\\ shape \end{tabular}}     & GVF-0.4-CCW  & \textbf{2.3501}                                                                        & 0.3129                                                                          & \textbf{0.2221}                                                                                               & \textbf{0.1817}                                                                                  & \textbf{0.0}                                                                           \\
                                                                              & MPC-0.4-CCW  & 0.7172                                                                                 & \textbf{0.3845}                                                                 & 0.6407                                                                                                        & 0.2131                                                                                           & 0.3894                                                                                 \\ 
\cline{2-7}
                                                                              & GVF-0.4-CW   & \textbf{2.3182}                                                                        & 0.3168                                                                          & \textbf{0.2317}                                                                                               & \textbf{0.2150}                                                                                  & \textbf{0.0006}                                                                        \\
                                                                              & MPC-0.4-CW   & 0.4324                                                                                 & \textbf{0.3905}                                                                 & 0.7662                                                                                                        & 0.2264                                                                                           & 0.5223                                                                                 \\ 
\cline{2-7}
                                                                              & GVF-0.25-CCW & \textbf{1.9326}                                                                        & \textbf{0.2472}                                                                 & 0.1890                                                                                                        & \textbf{0.1509}                                                                                  & 0.0                                                                                    \\
                                                                              & MPC-0.25-CCW & 1.1559                                                                                 & 0.2435                                                                          & \textbf{0.1664}                                                                                               & 0.1720                                                                                           & 0.0                                                                                    \\
\hline
\end{tabular}
\end{table}

\begin{table}
\centering
\caption{Comparison of comfort of GVF method and MPC on all the test roads at different speeds and directions}
\label{table_jackal_result_gvf_vs_mpc2}
\begin{tabular}{|l|l|l|l|l|l|} 
\hline
                                                                              & Experiment & \multicolumn{1}{c|}{\begin{tabular}[c]{@{}c@{}}First \\ order\\ speed\\ comfort \textuparrow \end{tabular}} & \multicolumn{1}{c|}{\begin{tabular}[c]{@{}c@{}}Second \\ order\\ speed \\ comfort \textuparrow \end{tabular}} & \multicolumn{1}{c|}{\begin{tabular}[c]{@{}c@{}}First\\ order\\ steering\\ comfort \textuparrow \end{tabular}} & \multicolumn{1}{c|}{\begin{tabular}[c]{@{}c@{}}Second\\ order\\ steering \\ comfort \textuparrow \end{tabular}}  \\ 
\hline
\multirow{6}{*}{\begin{tabular}[c]{@{}l@{}}Rectangle\\ shape \end{tabular}}   & GVF-0.4-CCW  & \textbf{-0.0356}                                                                                 & \textbf{-0.2532}                                                                                   & \textbf{-0.2251}                                                                                   & \textbf{-1.3403}                                                                                      \\
                                                                              & MPC-0.4-CCW  & -0.0832                                                                                          & -0.7605                                                                                            & -1.2542                                                                                            & -8.1963                                                                                               \\ 
\cline{2-6}
                                                                              & GVF-0.4-CW   & \textbf{-0.0311}                                                                                 & \textbf{-0.2149}                                                                                   & \textbf{-0.1995}                                                                                   & \textbf{-1.1850}                                                                                      \\
                                                                              & MPC-0.4-CW   & -0.0944                                                                                          & -0.8916                                                                                            & -1.4328                                                                                            & -10.9570                                                                                              \\ 
\cline{2-6}
                                                                              & GVF-0.25-CCW & \textbf{-0.0009}                                                                                 & \textbf{-0.0112}                                                                                   & \textbf{-0.1466}                                                                                   & \textbf{-0.8890}                                                                                      \\
                                                                              & MPC-0.25-CCW & -0.0570                                                                                          & -0.5272                                                                                            & -0.6384                                                                                            & -3.5208                                                                                               \\ 
\hline
\multirow{6}{*}{\begin{tabular}[c]{@{}l@{}}Oval\\ shape \end{tabular}}        & GVF-0.4-CCW  & \textbf{-0.0348}                                                                                 & \textbf{-0.2423}                                                                                   & \textbf{-0.2191}                                                                                   & \textbf{-1.4632}                                                                                      \\
                                                                              & MPC-0.4-CCW  & -0.1026                                                                                          & -0.9301                                                                                            & -1.4119                                                                                            & -8.9051                                                                                               \\ 
\cline{2-6}
                                                                              & GVF-0.4-CW   & \textbf{-0.0241}                                                                                 & \textbf{-0.1638}                                                                                   & \textbf{-0.1674}                                                                                   & \textbf{-1.1451}                                                                                      \\
                                                                              & MPC-0.4-CW   & -0.0847                                                                                          & -0.7534                                                                                            & -1.3957                                                                                            & -9.0432                                                                                               \\ 
\cline{2-6}
                                                                              & GVF-0.25-CCW & \textbf{-0.0005}                                                                                 & \textbf{-0.0061}                                                                                   & \textbf{-0.0969}                                                                                   & \textbf{-0.7614}                                                                                      \\
                                                                              & MPC-0.25-CCW & -0.0657                                                                                          & -0.6273                                                                                            & -0.4830                                                                                            & -3.0566                                                                                               \\ 
\hline
\multirow{6}{*}{\begin{tabular}[c]{@{}l@{}}Complex\\ shape \end{tabular}}     & GVF-0.4-CCW  & \textbf{-0.0341}                                                                                 & \textbf{-0.2540}                                                                                   & \textbf{-0.2272}                                                                                   & \textbf{-1.5306}                                                                                      \\
                                                                              & MPC-0.4-CCW  & -0.0625                                                                                          & -0.5846                                                                                            & -1.2133                                                                                            & -8.1747                                                                                               \\ 
\cline{2-6}
                                                                              & GVF-0.4-CW   & \textbf{-0.0348}                                                                                 & \textbf{-0.2339}                                                                                   & \textbf{-0.2240}                                                                                   & \textbf{-1.3911}                                                                                      \\
                                                                              & MPC-0.4-CW   & -0.0809                                                                                          & -0.7521                                                                                            & -1.2861                                                                                            & -8.7905                                                                                               \\ 
\cline{2-6}
                                                                              & GVF-0.25-CCW & \textbf{-0.0006}                                                                                 & \textbf{-0.0082}                                                                                   & \textbf{-0.1696}                                                                                   & \textbf{-1.0394}                                                                                      \\
                                                                              & MPC-0.25-CCW & -0.0525                                                                                          & -0.4932                                                                                            & -0.6457                                                                                            & -3.6786                                                                                               \\
\hline
\end{tabular}
\end{table}

\begin{table}
\centering
\caption{Evaluation of the robustness of GVF method on damaged lane markings on all the test roads}
\label{table_jackal_result_N_vs_D1}
\begin{tabular}{|l|l|l|l|l|l|l|} 
\hline
                                                                                & Experiment & \multicolumn{1}{c|}{\begin{tabular}[c]{@{}c@{}}Reward\\ per \\ second \textuparrow \end{tabular}} & \multicolumn{1}{c|}{\begin{tabular}[c]{@{}c@{}}Average\\ speed \textuparrow \end{tabular}} & \multicolumn{1}{c|}{\begin{tabular}[c]{@{}c@{}}Average \\ offcentered\\ -ness \textdownarrow\\ (normalized) \end{tabular}} & \multicolumn{1}{c|}{\begin{tabular}[c]{@{}c@{}}Average \\ absolute\\ road angle \textdownarrow \end{tabular}} & \multicolumn{1}{c|}{\begin{tabular}[c]{@{}c@{}}Near \\ out\\ of lane \textdownarrow \end{tabular}}  \\ 
\hline
\multirow{2}{*}{\begin{tabular}[c]{@{}l@{}}Rectangle\\ shape \end{tabular}}     & GVF-0.4-CCW   & 2.6835                                                                                 & 0.3205                                                                          & \textbf{0.1345 }                                                                                              & \textbf{0.1315 }                                                                                 & 0.0                                                                                    \\
                                                                                & GVF-0.4-CCW-D & \textbf{2.7407 }                                                                       & \textbf{0.3261 }                                                                & 0.1358                                                                                                        & 0.1351                                                                                           & 0.0                                                                                    \\ 
\hline
\multirow{2}{*}{\begin{tabular}[c]{@{}l@{}}Oval\\ shape \end{tabular}}          & GVF-0.4-CCW   & \textbf{2.4046 }                                                                       & \textbf{0.3501 }                                                                & \textbf{0.2754 }                                                                                              & 0.2125                                                                                           & \textbf{0.0145 }                                                                       \\
                                                                                & GVF-0.4-CCW-D & 2.0728                                                                                 & 0.3279                                                                          & 0.3285                                                                                                        & \textbf{0.2089 }                                                                                 & 0.0719                                                                                 \\ 
\hline
\multirow{2}{*}{\begin{tabular}[c]{@{}l@{}} Complex\\ shape \end{tabular}}      & GVF-0.4-CCW   & \textbf{2.3501 }                                                                       & 0.3129                                                                          & \textbf{0.2221 }                                                                                              & \textbf{0.1817 }                                                                                 & \textbf{0.0 }                                                                          \\
                                                                                & GVF-0.4-CCW-D & 2.1059                                                                                 & \textbf{0.3284 }                                                                & 0.3125                                                                                                        & 0.2365                                                                                           & 0.0942                                                                                 \\
\hline
\end{tabular}
\end{table}

\begin{table}
\centering
\caption{Evaluation of the comfort of GVF method on damaged lane markings on all the test roads}
\label{table_jackal_result_N_vs_D2}
\begin{tabular}{|l|l|l|l|l|l|} 
\hline
                                                                                & Experiment & \multicolumn{1}{c|}{\begin{tabular}[c]{@{}c@{}}First \\ order\\ speed\\ comfort \textuparrow \end{tabular}} & \multicolumn{1}{c|}{\begin{tabular}[c]{@{}c@{}}Second \\ order\\ speed \\ comfort \textuparrow \end{tabular}} & \multicolumn{1}{c|}{\begin{tabular}[c]{@{}c@{}}First\\ order\\ steering\\ comfort \textuparrow \end{tabular}} & \multicolumn{1}{c|}{\begin{tabular}[c]{@{}c@{}}Second\\ order\\ steering \\ comfort \textuparrow \end{tabular}}  \\ 
\hline
\multirow{2}{*}{\begin{tabular}[c]{@{}l@{}}Rectangle\\ shape \end{tabular}}     & GVF-0.4-CCW   & \textbf{-0.0356 }                                                                                & \textbf{-0.2532 }                                                                                  & \textbf{-0.2251 }                                                                                  & \textbf{-1.3403 }                                                                                     \\
                                                                                & GVF-0.4-CCW-D & -0.0383                                                                                          & -0.2715                                                                                            & -0.2303                                                                                            & -1.4620                                                                                               \\ 
\hline
\multirow{2}{*}{\begin{tabular}[c]{@{}l@{}}Oval\\ shape \end{tabular}}          & GVF-0.4-CCW   & -0.0348                                                                                          & \textbf{-0.2423 }                                                                                  & -0.2191                                                                                            & \textbf{-1.4632 }                                                                                     \\
                                                                                & GVF-0.4-CCW-D & \textbf{-0.0334 }                                                                                & -0.2953                                                                                            & \textbf{-0.2094 }                                                                                  & -1.6612                                                                                               \\ 
\hline
\multirow{2}{*}{\begin{tabular}[c]{@{}l@{}} Complex\\ shape \end{tabular}}      & GVF-0.4-CCW   & \textbf{-0.0341 }                                                                                & \textbf{-0.2540 }                                                                                  & \textbf{-0.2272 }                                                                                  & \textbf{-1.5306 }                                                                                     \\
                                                                                & GVF-0.4-CCW-D & -0.0437                                                                                          & -0.3608                                                                                            & -0.2897                                                                                            & -2.0946                                                                                               \\
\hline
\end{tabular}
\end{table}

\begin{table}
\centering
\caption{Comparison of GVF-BCQ and E2E-BCQ on the Rectangle test road at 0.4 m/s}
\label{table_jackal_result_gvf_vs_e2e1}
\begin{tabular}{|l|l|l|l|l|l|l|} 
\hline
                                                                                  & Experiment & \multicolumn{1}{c|}{\begin{tabular}[c]{@{}c@{}}Reward\\ per \\ second \textuparrow \end{tabular}} & \multicolumn{1}{c|}{\begin{tabular}[c]{@{}c@{}}Average\\ speed \textuparrow \end{tabular}} & \multicolumn{1}{c|}{\begin{tabular}[c]{@{}c@{}}Average \\ offcentered\\ -ness \textdownarrow\\ (normalized) \end{tabular}} & \multicolumn{1}{c|}{\begin{tabular}[c]{@{}c@{}}Average \\ absolute\\ road angle \textdownarrow \end{tabular}} & \multicolumn{1}{c|}{\begin{tabular}[c]{@{}c@{}}Near \\ out\\ of lane \textdownarrow \end{tabular}}  \\ 
\hline
\multirow{4}{*}{\begin{tabular}[c]{@{}l@{}}Rectangle\\ shape \end{tabular}}       & GVF-0.4-CCW                 & \textbf{2.6835 }                                                                       & \textbf{0.3205 }                                                                & \textbf{0.1345 }                                                                                              & \textbf{0.1315 }                                                                                 & \textbf{0.0 }                                                                          \\
                                                                                  & E2E-0.4-CCW                 & 1.2578                                                                                 & 0.1816                                                                          & 0.2558                                                                                                        & 0.2414                                                                                           & 0.0376                                                                                 \\ 
\cline{2-7}
                                                                                  & GVF-0.4-CW                  & \textbf{2.2915}                                                                        & \textbf{0.3140}                                                                 & \textbf{0.2217}                                                                                               & \textbf{0.1586}                                                                                  & \textbf{0.0}                                                                           \\
                                                                                  & E2E-0.4-CW & -0.1302                                                                                & 0.1710                                                                          & 0.9927                                                                                                        & 0.3034                                                                                           & 0.5418                                                                                 \\
\hline
\end{tabular}
\end{table}

\begin{table}
\centering
\caption{Comparison of GVF-BCQ and E2E-BCQ comfort levels on the Rectangle test road at 0.4 m/s}
\label{table_jackal_result_gvf_vs_e2e2}
\begin{tabular}{|l|l|l|l|l|l|} 
\hline
                                                                                  & Experiment & \multicolumn{1}{c|}{\begin{tabular}[c]{@{}c@{}}First \\ order\\ speed\\ comfort \textuparrow \end{tabular}} & \multicolumn{1}{c|}{\begin{tabular}[c]{@{}c@{}}Second \\ order\\ speed \\ comfort \textuparrow \end{tabular}} & \multicolumn{1}{c|}{\begin{tabular}[c]{@{}c@{}}First\\ order\\ steering\\ comfort \textuparrow \end{tabular}} & \multicolumn{1}{c|}{\begin{tabular}[c]{@{}c@{}}Second\\ order\\ steering \\ comfort \textuparrow \end{tabular}}  \\ 
\hline
\multirow{4}{*}{\begin{tabular}[c]{@{}l@{}}Rectangle\\ shape \end{tabular}}       & GVF-0.4-CCW                 & \textbf{-0.0356 }                                                                                & \textbf{-0.2532 }                                                                                  & \textbf{-0.2251 }                                                                                  & \textbf{-1.3403 }                                                                                     \\
                                                                                  & E2E-0.4-CCW                 & -0.0154                                                                                          & -0.2266                                                                                            & -0.1109                                                                                            & -1.4240                                                                                               \\ 
\cline{2-6}
                                                                                  & GVF-0.4-CW                  & \textbf{-0.0311}                                                                                 & \textbf{-0.2149}                                                                                   & \textbf{-0.1995}                                                                                   & \textbf{-1.1850}                                                                                      \\
                                                                                  & E2E-0.4-CW & -0.0148                                                                                          & -0.1937                                                                                            & -0.1174                                                                                            & -1.3514                                                                                               \\
\hline
\end{tabular}
\end{table}

\end{document}